\documentclass[11pt]{article}

\usepackage[final]{acl}

\usepackage{times}
\usepackage{latexsym}

\usepackage[T1]{fontenc}

\usepackage[utf8]{inputenc}

\usepackage{microtype}

\usepackage{inconsolata}

\usepackage{graphicx}

\usepackage{times}
\usepackage{latexsym}

\usepackage{graphicx}
\usepackage{subfig}
\usepackage{cuted}
\usepackage{caption}

\usepackage{booktabs}
\usepackage{multirow}
\usepackage{longtable}

\usepackage{amsmath}
\usepackage{amssymb}
\usepackage{amsthm}
\usepackage{mathtools}

\usepackage{algorithm}
\usepackage{algpseudocode}

\usepackage{natbib}

\usepackage{xcolor}
\usepackage[normalem]{ulem}
\usepackage{hyperref}
\usepackage{cleveref}

\newcommand{\llava}{\textsc{LLaVA}}
\newcommand{\minigpt}{\textsc{MiniGPT4}}
\newcommand{\ft}{\textsc{FT}}
\newcommand{\liveedit}{\textsc{LiveEdit}}

\newcommand{\vice}{\textsc{VICE}}

\newcommand{\nq}{\textsc{NQ}}

\newcommand{\wbedit}{\textsc{WaterBird-Edit}}
\newcommand{\cub}{\textsc{CUB-200-2011}}
\newcommand{\evqa}{\textsc{EVQA}}
\newcommand{\eic}{\textsc{EIC}}
\newcommand{\vlkeb}{\textsc{VLKEB}}

\title{Diagnosing and Mitigating Evidence Misgrounding in Multimodal Knowledge Editing}

\author{\textbf{Zhiwu Wang\textsuperscript{1}},
        \textbf{Yichen Wu\textsuperscript{2,*}},
        \textbf{Jinqian Yang\textsuperscript{3}},
        \textbf{Yizhou Yang\textsuperscript{1}},
        \\
        \textbf{Renzhen Wang\textsuperscript{1}},
        \textbf{Qian Zhao\textsuperscript{1,*}},
        \textbf{Deyu Meng\textsuperscript{4,5}}
        \\
        \textsuperscript{1}School of Mathematics and Statistics, Xi'an Jiaotong University, Xi'an, Shaanxi, China
        \\
        \textsuperscript{2}Harvard Medical School, Boston, MA 02155, U.S.
        \\
        \textsuperscript{3}School of Mathematics and Statistics, Xidian University, Xi'an, Shaanxi, China
        \\
        \textsuperscript{4}School of Mathematics and Statistics and Ministry of Education Key Lab of Intelligent\\Networks and Network Security, Xi'an Jiaotong University, Xi'an, Shaanxi, China
        \\
        \textsuperscript{5}Macao Institute of Systems Engineering, Macau University of Science and Technology,\\Taipa, Macao, China
        \\
        \texttt{\{wangzhiwu7,wuyichen.am97,yyzh201314\}@gmail.com},
        \texttt{jinqianyang@stu.xidian.edu.cn},
        \\
        \texttt{\{rzwang,timmy.zhaoqian\}@xjtu.edu.cn},
        \texttt{dymeng@mail.xjtu.edu.cn}
}

\begin{document}

\maketitle

\begingroup
\renewcommand{\thefootnote}{*}
\footnotetext{Corresponding author.}
\endgroup


\begin{abstract}
Multimodal knowledge editing aims to update visually grounded facts in Multimodal Large Language Models (MLLMs) while preserving unrelated text and visual behaviors. However, standard answer-level metrics can hide \textit{evidence misgrounding}: an edit may produce the target answer on the original image yet remain sensitive to incidental context rather than the intended visual evidence.
To examine this issue in a controlled setting, we construct \wbedit{}, an analysis set that decouples bird habitat labels from background context and measures whether edited predictions remain stable when the target evidence is preserved but the background changes.
Motivated by this diagnosis, we propose \vice{} (\textbf{V}isual \textbf{I}n-\textbf{C}ontext \textbf{E}diting), a retrieval-based framework that stores edits in external memory and applies them through visual in-context demonstrations designed to expose task-relevant evidence.
Experiments on three standard benchmarks show that \vice{} achieves strong lifelong editing performance, while results on \wbedit{} show that it substantially reduces sensitivity to the controlled background shifts.
\footnote{Code and dataset are available in \url{https://github.com/Wang-Z-W/VICE-main}.}
\end{abstract}

\section{Introduction}
\label{sec:intro}

Multimodal Large Language Models (MLLMs) have become the backbone of modern vision-language understanding, encoding extensive factual knowledge about the visual world~\citep{li2023blip-2,zhu2024minigpt-4,liu2023llava,liu2024llava-v1.5,Qwen2-VL,Qwen3-VL}. When this knowledge becomes outdated or incorrect, multimodal knowledge editing~\citep{cheng2023mmedit,huang2024vlkeb,chen2025liveedit,shi2025dualedit} aims to update a visually grounded fact while preserving behavior on unrelated text and visual queries. Unlike text-only editing, a successful MLLM edit must apply the new fact according to the intended visual evidence rather than incidental image context.

\begin{figure}[t]
\centering
\includegraphics[width=\columnwidth]{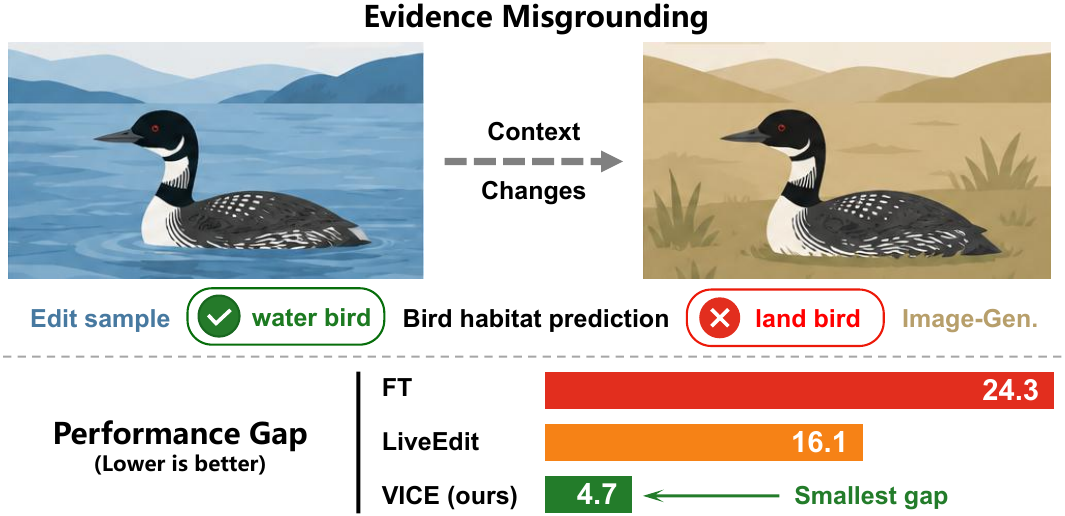}
\caption{An example of background-induced evidence misgrounding. An edit may succeed on the original image but fail when the same bird evidence appears with a different background. Lower performance gap indicates less sensitivity to background context.}
\label{fig:teaser}
\end{figure}

However, answer-level success does not guarantee visually grounded editing. In our diagnosis, an editor can produce the corrected answer on the original edit image but become unreliable on image variants that preserve the target evidence while changing incidental context. As illustrated by the bird-habitat example in \Cref{fig:teaser}, an edit may succeed on the original water-bird image, yet its prediction can still vary when similar bird evidence appears with a different background. We use the term \emph{evidence misgrounding} to describe this oberavble mismatch: an edit is successful at the answer level, yet its prediction remains sensitive to incidental visual context rather than being stable under changes that preserve the intended evidence. Because standard reliability, generality, and locality metrics mainly measure answer correctness, they can overestimate edit quality by overlooking context-sensitive image generalization.

To study this failure in a controlled setting, we use \wbedit{}, a diagnostic analysis inspired by \citet{sagawa2020group-dro}, where the edited fact is the bird's habitat label and the background provides correlated but potentially spurious context.
By decoupling bird habitat and background into four target-context groups, \wbedit{} tests whether an edited prediction tracks the bird evidence consistently across background changes. On this diagnosis set, we find that \ft{} and \liveedit{}~\citep{chen2025liveedit} exhibit behavior consistent with evidence misgrounding. Although these models increase the likelihood of the desired answer through parameter updates or hidden-state modulation, their predictions nevertheless remain sensitive to background changes.

These findings suggest that robust MLLM editing should make the evidence used to apply an edit explicit, rather than relying solely on implicit parameter or hidden-state updates. Motivated by this principle, we propose \vice{} (\textbf{V}isual \textbf{I}n-\textbf{C}ontext \textbf{E}diting), a lifelong framework that applies edits through visual in-context learning~\citep{zhou2024vicl} while freezing the backbone. For each edit, \vice{} constructs evidence-summary demonstrations whose image summaries highlight task-relevant evidence and separate it from incidental context, then stores the edited fact and its demonstrations in a memory. At inference time, a query retrieves the nearest edit from memory, and a learned scope head decides whether to answer with the retrieved edit context or bypass the edit. This allows the frozen MLLM to apply edits through demonstrations designed to emphasize task-relevant evidence while preserving unrelated behavior.

Our contributions are threefold. First, we study evidence misgrounding in MLLM editing through \wbedit{}, a controlled diagnostic setting that measures whether edited behavior remains sensitive to incidental context after controlling for the target evidence. Second, we propose \vice{}, a frozen-backbone framework that applies edits through visual in-context learning. Third, experiments on EVQA~\citep{cheng2023mmedit}, EIC~\citep{cheng2023mmedit}, VLKEB~\citep{huang2024vlkeb}, and \wbedit{} demonstrate strong lifelong editing performance and reduced sensitivity to the controlled background shifts.

\section{Diagnosing Evidence Misgrounding}
\label{sec:diagnosis}

We use evidence misgrounding to denote the case where an edited model yields the corrected answer on the edit image while remaining overly sensitive to incidental visual context rather than consistently following the intended visual evidence. Evidence misgrounding harms edited-model generalization, so detecting this failure can help build robust editing mechanisms. However, existing multimodal editing benchmarks make diagnosis difficult: target objects, scenes, and other visual cues are usually entangled, obscuring whether an edited answer follows the intended evidence or a correlated cue. We therefore construct a controlled diagnostic set with explicit target-context groups to test whether edited behavior follows target evidence across contexts.

\begin{figure*}[t]
\centering
\includegraphics[width=\textwidth]{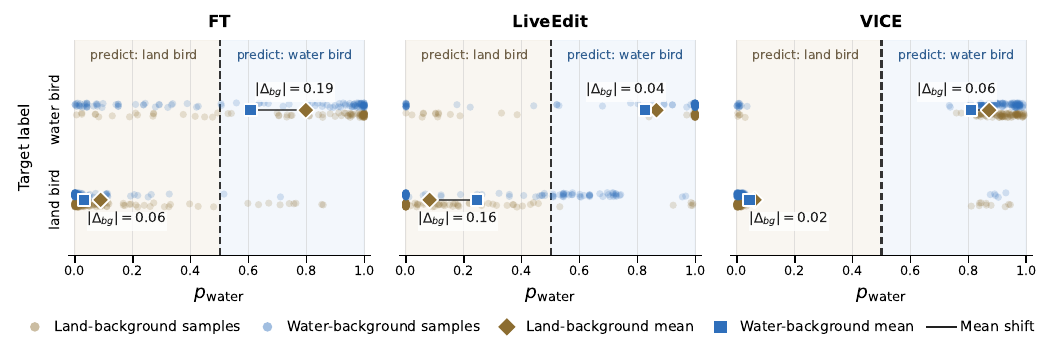}
\caption{
Sample-wise $p_{\text{water}}$ defined in~\Cref{eq:p_water} for \wbedit{} image-rephrase samples after editing.
The top two bands show water-bird targets, and the bottom two bands show land-bird targets.
Point colors indicate background labels: blue for water backgrounds and orange for land backgrounds.
Large markers denote background-group means, and $\Delta_{bg}$ reports the background-induced mean shift for each target label.
}
\label{fig:prob_distribution}
\end{figure*}

\subsection{\wbedit{} Diagnostic Analysis}

To make evidence misgrounding observable, we construct \wbedit{},
a controlled diagnostic analysis set built on \cub{}~\citep{wah2011cub} for lifelong editing, inspired by \citet{sagawa2020group-dro}.

We first formulate a binary VQA task: given a CUB image, a prompt such as \texttt{``What type of bird is in the image? Choose from: land, water.''},
the model predicts the species-level bird habitat label $y \in \{\text{water}, \text{land}\}$.
Each image is also assigned a background label $b \in \{\text{water}, \text{land}\}$ by a pretrained vision-language model.
The habitat and background labels then form four target-context groups $(y,b)\in\{\text{water},\text{land}\}^2$.
In this setting, evidence misgrounding is diagnosed when edited predictions vary with $b$ even though the desired answer is determined by $y$.
To construct the editing samples, we first query the MLLM to be edited on the binary habitat VQA task and identify images where its habitat prediction is incorrect, resulting in 500 such images. Each retained image is then used to form an editing instance, where the target answer is set to the ground-truth habitat label to correct the model's original prediction.
The filtered editing samples remain imbalanced across the four target-context groups, thereby preserving the correlation between habitat labels and background labels that the editor may exploit during editing.
For the Image-Generality evaluation, each editing sample is paired with another image of the same bird species, which we call image-rephrase sample. This paired image shares the same habitat label as the original editing image but may have a different background, allowing us to evaluate whether the edit generalizes across images of the same species.
This design tests whether an editor exposed to spuriously correlated visual cues still generalizes well according to the target bird rather than the background.
Details are in Appendix \ref{sec:appendix_wbedit_construction}.

After the 500-edit sequence, we evaluate each editor on the image-rephrase samples and report:
(1) I-Gen., the standard Image-Generality metric,
(2) Group Image-Generality $\text{IG}(g)$, a fine-grained I-Gen. within each group $g=(y,b)$,
(3) Performance Gap (PG), defined as $\text{I-Gen.}-\min_g\text{IG}(g)$, and
(4) sample-wise normalized probability assigned to the ``water'' answer among the two candidate answers (``water'' and ``land''):
\begin{equation}
\label{eq:p_water}
p_{\text{water}}=\frac{\exp(s_{\text{water}})}{\exp(s_{\text{water}})+\exp(s_{\text{land}})},
\end{equation}
where $s_{\text{water}}$ and $s_{\text{land}}$ are the model logits for the two candidate answers.
A large PG indicates target-context sensitivity and functions as a worst-group robustness gap,
while the distribution of $p_{\text{water}}$ across groups shows how this sensitivity appears at the sample level.
We further construct object-only image-rephrase variants using the segmentation masks provided by \cub{},
and report I-Gen. on these variants to measure how edited predictions change after removing the background while preserving the target bird evidence.

\begin{table}[t]
\centering
\small
\setlength{\tabcolsep}{2.8pt}
\resizebox{\columnwidth}{!}{
\begin{tabular}{lccccccc}
\toprule
\multirow{2}{*}{\textbf{Method}}
& \multirow{2}{*}{\textbf{I-Gen.}$\uparrow$}
& \multirow{2}{*}{\textbf{Mask I-Gen.}$\uparrow$}
& \multicolumn{4}{c}{\textbf{Group I-Gen.}$\uparrow$}
& \multirow{2}{*}{\textbf{PG}$\downarrow$} \\
\cmidrule(lr){4-7}
& & & WW & WL & LW & LL & \\
\midrule
\ft & 85.2 & 87.8 & 60.9 & 80.6 & 98.5 & 94.3 & 24.3 \\
\liveedit & 83.8 & 92.4 & 84.3 & 84.9 & 67.7 & 96.2 & 16.1 \\
\vice & 93.4 & 89.0 & 88.7 & 94.6 & 96.2 & 93.7 & 4.7 \\
\bottomrule
\end{tabular}
}
\caption{\wbedit{} diagnostics on image-rephrase samples.
Each group is written as (habitat, background), where ``W'' and ``L'' denote water and land.
``PG'' denotes performance gap.
``Mask I-Gen.'' is measured on object-only image-rephrase variants that preserve the bird region and mask the background.}
\label{tab:p4_results}
\end{table}

\subsection{Context Sensitivity After Editing}

We focus on two MLLM editors: \ft{}, a direct parameter-update baseline, and \liveedit{}~\citep{chen2025liveedit}, a recent lifelong MLLM editing method that modulates hidden states. The diagnostic experiments lead to the following findings.

\noindent\textbf{Answer-level success can hide target-context sensitivity.}
As shown in \Cref{tab:p4_results}, \ft{} and \liveedit{} achieve reasonable full-set I-Gen. scores of $85.2$ and $83.8$,
yet their worst groups are much lower, producing large PG values of $24.3$ and $16.1$.
This gap shows that their apparent image generalization is not robust under changes in the incidental background context.
\Cref{fig:prob_distribution} shows the same pattern at the probability level: for a fixed target label,
changing the background substantially shifts $p_{\text{water}}$,
especially for \ft{} on water targets ($|\Delta_{bg}|=0.19$) and \liveedit{} on land targets ($|\Delta_{bg}|=0.16$).

\noindent\textbf{Object-only masking provides a complementary background intervention.}
Object-only masking provides a complementary diagnostic by preserving the target bird region while removing the surrounding background.
\Cref{tab:p4_results} shows that
masking improves \ft{} from $85.2$ to $87.8$ and \liveedit{} from $83.8$ to $92.4$,
indicating that background context harms their original image-generality performance.

\begin{figure*}[t]
    \centering
    \includegraphics[width=\linewidth]{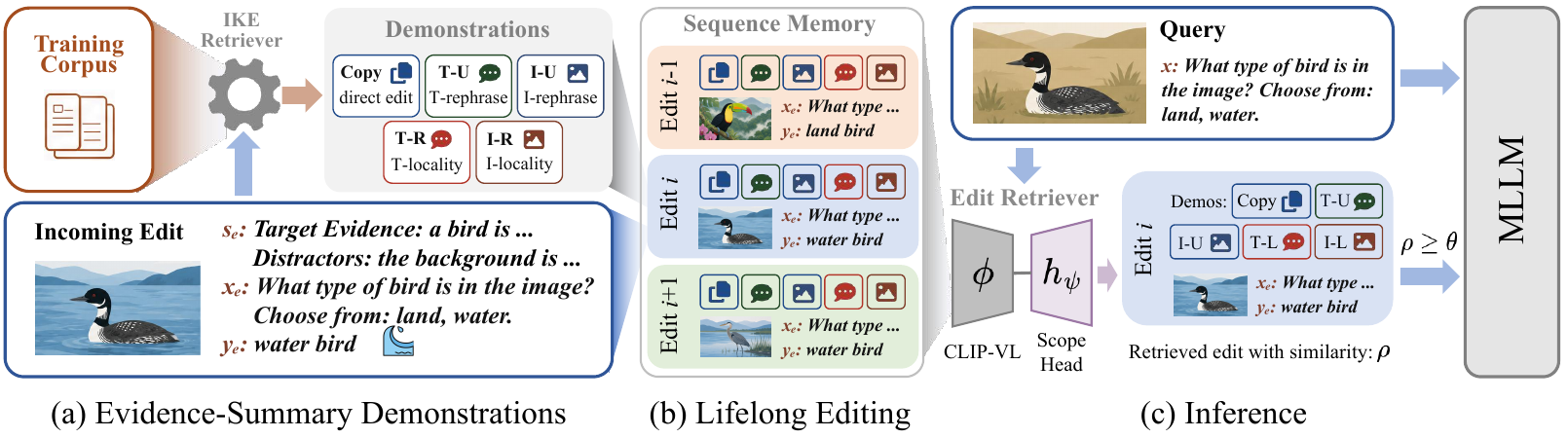}
    \caption{Illustration of the \vice{} pipeline. (a) Evidence-summary demonstrations for Copy, Text-Update (T-U), Image-Update (I-U), Text-Retain (T-R), and Image-Retain (I-R), all using the same template in~\Cref{eq:vice_demo_template}. (b) Lifelong editing, where each incoming new fact is stored together with its summary and retrieved demonstrations in Sequence Memory. (c) Inference, where a multimodal encoder retrieves the most relevant edit and routes the query augmented by the retrieved context to the MLLM when the retrieval similarity score exceeds the threshold.}
    \label{fig:vice_pipeline}
\end{figure*}

Together, these diagnostics show that methods based on parameter updates or hidden-state modulation can achieve strong answer-level editing performance while their edited predictions remain sensitive to background context.
This observation motivates \vice{}, which keeps the MLLM backbone frozen and applies edits through visual in-context demonstrations that explicitly represent task-relevant evidence and incidental cues.

\section{Method: VICE}
\label{sec:method}

Building on the above diagnosis, we propose \vice{}, a frozen-backbone framework that applies edits through retrieved visual demonstrations at inference time.
Each edit is stored with evidence-summary demonstrations that highlight task-relevant visual cues and separate them from incidental context.
Given a query, \vice{} retrieves the most relevant edit and uses scope-aware routing to decide whether the edit should be applied. \Cref{fig:vice_pipeline} illustrates the \vice{} pipeline.

\subsection{Preliminary: In-Context Editing}
\citet{zheng2023ike} propose IKE and frame editing as prompting with demonstrations rather than updating parameters.
Given a new fact $(x_e,y_e)$ and a query $x$, IKE forms a test-time prompt as
$``\mathcal C.\ \text{New Fact: }x_e\ y_e.\ \text{Prompt: }x"$, where $\mathcal C$ contains retrieved demonstrations that teach when to copy, update, or retain answers.
Each demonstration $c_k\in\mathcal C$ uses the same two textual fields:
\begin{equation}
c_k = \text{``New Fact: }x_e^k\ y_e^k\text{. Prompt: }\tilde{x}^k\ \tilde{y}^k\text{''},
\end{equation}
where both the demonstration fact $(x_e^k,y_e^k)$ and the prompt-answer pair $(\tilde{x}^k,\tilde{y}^k)$ are drawn from training-set examples.
IKE retrieves these demonstrations using sentence-level similarity between the requested new fact and candidate training entries.
We follow this retrieval-based protocol for selecting demonstration content.
For MLLM editing, \vice{} keeps the ``New Fact / Prompt'' template but adds an ``Image Summary'' field, introduced next, so that demonstrations can expose textualized visual evidence for the demonstrated fact.

\begin{table}[t]
    \centering
    \resizebox{\columnwidth}{!}{
    \begin{tabular}{cccc}
    \toprule
      Type & Image Summary & New Fact & Prompt\\
    \midrule
      Copy & $s_e=G(i_e,x_e)$ & $(x_e,y_e)$ & $(x_e,y_e)$\\
      Text-Update & $s_e=G(i_e,x_e)$ & $(x_e,y_e)$ & $(x_r,y_e)$\\
      Image-Update & $s_r=G(i_r,x_e)$ & $(x_e,y_e)$ & $(x_e,y_e)$\\
      Text-Retain & $s_e=G(i_e,x_e)$ & $(x_e,y_e)$ & $(x_{l},y_{l})$\\
      Image-Retain & $s_{l}=G(i_l,x_{il})$ & $(x_e,y_e)$ & $(x_{il},y_{il})$\\
    \bottomrule
    \end{tabular}
    }
    \caption{Five evidence-summary demonstration roles instantiated from the template in \Cref{eq:vice_demo_template}. $(i_e,x_e,y_e)$ is the anchor example; $x_r$ is a text rephrase with answer $y_e$; $i_r$ is an image variant for prompt $x_e$; $(x_l,y_l)$ is a text-locality example; and $(i_l,x_{il},y_{il})$ is an image-locality example. Superscript of the $k$-th demonstration $c_k$ is omitted for readability.}
    \label{tab:demo}
\end{table}

\subsection{Evidence-Summary Demonstrations}
\label{sec:context_pool}

Motivated by a recent visual in-context learning method~\citep{zhou2024vicl}, \vice{} extends IKE's textual demonstrations by adding an evidence-summary field for multimodal edits.
The goal is to make the visual evidence associated with the demonstrated fact explicit in the prompt, rather than leaving the ``New Fact'' as an answer-only association.
Starting from IKE's ``New Fact / Prompt'' demonstration, \vice{} inserts an ``Image Summary'' field before the ``New Fact'':
\begin{equation}
\begin{aligned}
c_k = \text{``}&\text{Image Summary: }s^k\text{.}\\
&\text{New Fact: }x_e^k\ y_e^k\text{. Prompt: }\tilde{x}^k\ \tilde{y}^k\text{''}.
\end{aligned}
\label{eq:vice_demo_template}
\end{equation}
This keeps the textual interface of in-context editing while allowing a demonstration to state the image evidence associated with its own ``New Fact'' field.
Concretely, given an edit image and prompt $(i_e^k, x_e^k)$, we use a frozen summary model $G$ to generate a text-guided image summary
\begin{equation}
s_e^k = G(i_e^k, x_e^k),
\end{equation}
where $s_e^k$ describes object evidence relevant to $x_e^k$ and separates it from incidental visual attributes.
The generation prompt explicitly instructs the summary model not to answer the question directly, so the summary remains evidence-focused rather than revealing or restating the edit target.
In our implementation, $G$ can be instantiated as the same pretrained MLLM before editing,
which avoids introducing a stronger external teacher. Please refer to Appendix \ref{sec:appendix_vice_details} for image-summary generation details and the full augmented-prompt template.

\Cref{tab:demo} summarizes five demonstration roles. ``Copy'' shows direct use of the ``New Fact''. ``Text-Update'' and ``Image-Update'' show transfer to a rephrased question or a related image, respectively. ``Text-Retain'' and ``Image-Retain'' pair the ``New Fact'' with unrelated queries whose original answers should be preserved.

\subsection{Lifelong Visual In-Context Editing}
In the lifelong setting, edits arrive sequentially.
For each edit $(i_e,x_e,y_e)$, \vice{} builds the image summary $s_e=G(i_e,x_e)$, retrieves the evidence-summary demonstrations $\mathcal C$ described in \Cref{sec:context_pool}, following IKE's sentence-level similarity retrieval protocol, and appends the resulting record to an external \textbf{Sequence Memory} $\mathcal M$:
\begin{equation}
\mathcal M \gets \mathcal M \cup \{(i_e, x_e, y_e, s_e, \mathcal{C})\}.
\end{equation}
Editing is therefore a simple memory-write operation and does not modify model parameters or hidden states.

At inference time, \vice{} decides whether a query should use a stored edit.
Retrieval first proposes a memory entry, and scope verification then decides whether to apply it.
We construct a retrieval encoder from a frozen CLIP-based multimodal encoder $\phi$ and a learnable scope-discriminative head $h_{\psi}$.
The encoder $\phi$ concatenates image and text features, while $h_{\psi}$ projects them into a new space that separates in-scope and out-of-scope knowledge.
Given a query $(i,x)$, \vice{} retrieves the most similar memory entry:
\begin{equation}
\label{equ:retrieve}
m^*=\mathop{\arg\max}\limits_{m\in\mathcal M}\mathbf{z}^{\top}\mathbf{z}_m,
\end{equation}
where $\mathbf{z}=h_{\psi}(\phi(i,x)),\mathbf{z}_m=h_{\psi}(\phi(i_m,x_m))$, $i_m,x_m$ are entries in a cached memory $m$, and
$m^* = (i_e^*, x_e^*, y_e^*, s_e^*, \mathcal{C}^*)$. We use $\rho=\max_{m\in\mathcal M}\mathbf{z}^{\top}\mathbf{z}_m$ as the retrieval score.
Given a fixed threshold $\theta$, if $\rho \geq \theta$, the retrieved edit is applied to form an augmented prompt:
\begin{equation}
\label{equ:aug_prompt}
\begin{aligned}
``\mathcal C^*.\ &\text{Image Summary: }s_e^*\text{.}\\
&\text{New Fact: }x_e^*\ y_e^*\text{. Prompt: }x\text{''},
\end{aligned}
\end{equation}
along with the original image input $i$ to query the MLLM.
Otherwise, the model receives only the original query without any retrieved edit information.
\Cref{alg:vice_inference} summarizes the inference.

\makeatletter
\renewcommand{\theHALG@line}{\thealgorithm.\arabic{ALG@line}}
\makeatother

\begin{algorithm}[t]
\caption{\vice{} Inference}
\label{alg:vice_inference}
\begin{algorithmic}
\Require Frozen MLLM, sequence memory $\mathcal{M}$, multimodal encoder $\phi$, scope-discriminative head $h_{\psi}$, threshold $\theta$, query $(i,x)$
\Ensure Prediction $\hat{y}$
\State $\mathbf{z} \gets h_{\psi}(\phi(i,x))$
\State $(i_e^*,x_e^*,y_e^*,s_e^*,\mathcal{C}^*), \rho \gets$ \Cref{equ:retrieve}
\If{$\rho \geq \theta$}
    \State $p \gets$ \Cref{equ:aug_prompt}
    \State $\hat{y} \gets \text{MLLM}(i,p)$
\Else
    \State $\hat{y} \gets \text{MLLM}(i,x)$
\EndIf
\State \Return $\hat{y}$
\end{algorithmic}
\end{algorithm}

\begin{table*}[t]
\centering
\footnotesize
\setlength{\tabcolsep}{1.2pt}
\newcommand{\vicecell}[1]{\cellcolor{blue!6}#1}
\resizebox{\textwidth}{!}{%
\begin{tabular}{lll*{15}{c}}
\toprule
& & & \multicolumn{5}{c}{EVQA} & \multicolumn{5}{c}{EIC} & \multicolumn{5}{c}{VLKEB} \\
\cmidrule(lr){4-8}\cmidrule(lr){9-13}\cmidrule(lr){14-18}
\textbf{Model} & \textbf{\# Edits} & \textbf{Method} & Rel & T-Gen & I-Gen & T-Loc & I-Loc & Rel & T-Gen & I-Gen & T-Loc & I-Loc & Rel & T-Gen & I-Gen & T-Loc & I-Loc \\
\midrule
\multirow{18}{*}{LLaVA-v1.5}
& \multirow{6}{*}{1}    & FT-VL    & 100.00 & \underline{91.03} & \underline{95.38} & 95.29 & 90.53 & \textbf{99.93} & \textbf{94.65} & \textbf{97.34} & 91.63 & 82.16 & \textbf{100.00} & \textbf{98.74} & \textbf{99.92} & 76.13 & 84.15 \\
&                       & TP       & 33.27 & 31.33 & 29.06 & \underline{96.77} & \underline{98.83} & 57.49 & 57.48 & 53.91 & 95.41 & \underline{96.52} & 50.34 & 52.43 & 50.27 & \underline{92.59} & 96.58 \\
&                       & MEND     & 91.23 & 90.05 & 91.29 & 91.02 & 90.22 & 92.82 & 91.81 & \underline{90.59} & \underline{96.38} & 93.69 & 92.13 & 91.28 & 90.22 & 89.19 & 90.13 \\
&                       & SERAC    & 89.33 & 83.72 & 84.97 & 82.05 & 23.78 & 88.18 & 81.03 & 85.61 & 84.01 & 28.58 & 89.77 & 89.11 & 87.92 & 66.68 & 14.20 \\
&                       & LiveEdit & \underline{94.28} & \textbf{94.51} & 88.01 & 100.00 & 100.00 & 82.16 & 81.01 & 78.27 & 100.00 & 100.00 & 96.43 & 95.22 & 93.72 & 100.00 & \textbf{100.00} \\
&                    & \vicecell{VICE} & \vicecell{\textbf{100.00}} & \vicecell{87.33} & \vicecell{\textbf{97.57}} & \vicecell{\textbf{100.00}} & \vicecell{\textbf{100.00}} & \vicecell{\underline{98.91}} & \vicecell{\underline{92.76}} & \vicecell{79.61} & \vicecell{\textbf{100.00}} & \vicecell{\textbf{100.00}} & \vicecell{\underline{99.98}} & \vicecell{\underline{98.27}} & \vicecell{\underline{96.72}} & \vicecell{\textbf{100.00}} & \vicecell{\underline{99.98}} \\
\cmidrule(lr){2-18}
& \multirow{6}{*}{100}  & FT-VL    & 76.44 & 67.30 & 67.76 & 58.04 & 46.18 & 73.11 & 69.78 & \underline{65.46} & 52.21 & 58.69 & 75.91 & 72.95 & 74.89 & 56.36 & 70.29 \\
&                       & TP       & 32.60 & 30.74 & 28.65 & \underline{83.13} & 89.68 & 50.31 & 50.67 & 46.83 & 34.35 & 63.78 & 48.26 & 50.78 & 48.20 & \underline{64.18} & 79.15 \\
&                       & MEND     & 2.22 & 2.20 & 2.21 & 0.21 & 0.62 & 56.83 & 56.89 & 53.07 & \underline{87.63} & 84.64 & 0.56 & 0.58 & 0.66 & 0.18 & 0.07 \\
&                       & SERAC    & 88.08 & 81.53 & 82.48 & 62.13 & 12.90 & 53.35 & 53.70 & 49.41 & 48.04 & 17.25 & 72.25 & 62.43 & 70.68 & 53.73 & 13.69 \\
&                       & LiveEdit & \underline{93.54} & \textbf{92.34} & \underline{85.89} & 100.00 & \underline{99.31} & \underline{80.81} & \underline{78.77} & 63.52 & 100.00 & \textbf{100.00} & \underline{94.56} & \underline{90.65} & \underline{89.56} & 100.00 & \textbf{100.00} \\
&                       & \vicecell{VICE} & \vicecell{\textbf{100.00}} & \vicecell{\underline{87.15}} & \vicecell{\textbf{97.50}} & \vicecell{\textbf{100.00}} & \vicecell{\textbf{99.82}} & \vicecell{\textbf{98.91}} & \vicecell{\textbf{92.76}} & \vicecell{\textbf{73.97}} & \vicecell{\textbf{100.00}} & \vicecell{\underline{99.95}} & \vicecell{\textbf{99.98}} & \vicecell{\textbf{98.27}} & \vicecell{\textbf{97.14}} & \vicecell{\textbf{100.00}} & \vicecell{\underline{99.85}} \\
\cmidrule(lr){2-18}
& \multirow{6}{*}{1000} & FT-VL    & 60.21 & 50.76 & 51.03 & 24.37 & 29.48 & 64.21 & 59.43 & \underline{57.09} & 33.39 & 38.30 & 65.30 & 62.67 & 63.87 & 33.89 & 37.32 \\
&                       & TP       & 32.82 & 30.48 & 28.55 & 42.12 & 60.28 & 33.75 & 36.27 & 32.34 & 10.47 & 26.22 & 34.23 & 34.74 & 34.44 & 27.11 & 45.52 \\
&                       & MEND     & 0.04 & 0.05 & 0.05 & 0.08 & 0.09 & 54.39 & 54.14 & 50.99 & \underline{83.87} & 80.60 & 0.03 & 0.05 & 0.07 & 0.06 & 0.08 \\
&                       & SERAC    & 85.57 & 75.58 & 82.01 & \underline{62.46} & 15.69 & 52.93 & 53.44 & 49.01 & 49.91 & 16.65 & 60.93 & 56.49 & 60.06 & \underline{52.94} & 15.04 \\
&                       & LiveEdit & \underline{92.93} & \textbf{90.16} & \underline{84.30} & 100.00 & \underline{96.43} & \underline{72.80} & \underline{69.95} & 57.05 & 100.00 & \underline{99.79} & \underline{92.22} & \underline{83.97} & \underline{82.75} & 100.00 & \textbf{100.00} \\
&                       & \vicecell{VICE} & \vicecell{\textbf{100.00}} & \vicecell{\underline{87.03}} & \vicecell{\textbf{97.01}} & \vicecell{\textbf{100.00}} & \vicecell{\textbf{98.38}} & \vicecell{\textbf{98.91}} & \vicecell{\textbf{92.76}} & \vicecell{\textbf{60.91}} & \vicecell{\textbf{100.00}} & \vicecell{\textbf{99.90}} & \vicecell{\textbf{99.98}} & \vicecell{\textbf{98.26}} & \vicecell{\textbf{96.75}} & \vicecell{\textbf{100.00}} & \vicecell{\underline{98.32}} \\
\midrule
\multirow{18}{*}{MiniGPT4}
& \multirow{6}{*}{1}    & FT-VL    & \textbf{100.00} & \textbf{99.74} & \textbf{99.86} & 63.66 & 35.65 & \textbf{100.00} & \textbf{95.28} & \textbf{96.41} & 58.21 & 33.84 & \textbf{100.00} & \textbf{99.75} & \textbf{99.95} & 56.12 & 42.03 \\
&                       & TP       & 30.07 & 28.55 & 30.40 & \underline{89.48} & 71.50 & 30.79 & 32.00 & 34.41 & 44.72 & 28.74 & 37.17 & 38.31 & 37.49 & 74.56 & 46.15 \\
&                       & MEND     & 93.47 & 91.32 & 91.22 & 81.51 & \underline{74.97} & 90.28 & 83.91 & 79.05 & \underline{95.92} & \underline{93.97} & 90.28 & 89.18 & 89.55 & \underline{93.63} & 86.69 \\
&                       & SERAC    & 87.66 & 69.24 & 85.96 & 82.11 & 24.11 & 85.09 & 78.96 & \underline{81.70} & 83.36 & 24.50 & 89.28 & 89.06 & 87.35 & 64.88 & 12.72 \\
&                       & LiveEdit & 94.90 & \underline{92.63} & 92.06 & 100.00 & 100.00 & 86.84 & 83.82 & 76.95 & 100.00 & 100.00 & \underline{96.41} & \underline{95.56} & \underline{92.85} & 100.00 & \textbf{100.00} \\
&                       & \vicecell{VICE} & \vicecell{\underline{99.98}} & \vicecell{88.74} & \vicecell{\underline{97.35}} & \vicecell{\textbf{100.00}} & \vicecell{\textbf{100.00}} & \vicecell{\underline{98.93}} & \vicecell{\underline{93.67}} & \vicecell{77.85} & \vicecell{\textbf{100.00}} & \vicecell{\textbf{100.00}} & \vicecell{95.27} & \vicecell{82.95} & \vicecell{92.05} & \vicecell{\textbf{100.00}} & \vicecell{\underline{99.99}} \\
\cmidrule(lr){2-18}
& \multirow{6}{*}{100}  & FT-VL    & 56.02 & 49.77 & 50.35 & 16.97 & 10.42 & 59.24 & 56.32 & 52.58 & 4.13 & 4.65 & 64.49 & 63.10 & 62.70 & 21.95 & 14.10 \\
&                       & TP       & 6.64 & 6.16 & 6.04 & 1.80 & 0.61 & 7.74 & 7.44 & 7.23 & 1.08 & 0.63 & 4.93 & 4.85 & 4.40 & 1.57 & 1.34 \\
&                       & MEND     & 23.32 & 20.75 & 21.42 & 47.27 & 42.60 & 51.92 & 53.96 & 48.24 & \underline{86.02} & 83.88 & 38.85 & 40.02 & 37.99 & \underline{72.82} & 68.89 \\
&                       & SERAC    & 87.39 & 66.53 & 81.70 & \underline{65.13} & 14.04 & 47.41 & 46.17 & 45.22 & 49.37 & 13.52 & 69.38 & 58.31 & 67.57 & 51.71 & 13.72 \\
&                       & LiveEdit & \underline{93.23} & \textbf{91.27} & \underline{88.34} & 100.00 & \underline{99.61} & \underline{84.38} & \underline{83.53} & \textbf{71.85} & 100.00 & \textbf{100.00} & \underline{94.06} & \textbf{94.62} & \underline{89.42} & 100.00 & \textbf{100.00} \\
&                       & \vicecell{VICE} & \vicecell{\textbf{99.98}} & \vicecell{\underline{88.49}} & \vicecell{\textbf{97.28}} & \vicecell{\textbf{100.00}} & \vicecell{\textbf{99.82}} & \vicecell{\textbf{98.93}} & \vicecell{\textbf{93.67}} & \vicecell{\underline{71.73}} & \vicecell{\textbf{100.00}} & \vicecell{\underline{99.95}} & \vicecell{\textbf{95.27}} & \vicecell{\underline{82.95}} & \vicecell{\textbf{92.44}} & \vicecell{\textbf{100.00}} & \vicecell{\underline{99.87}} \\
\cmidrule(lr){2-18}
& \multirow{6}{*}{1000} & FT-VL    & 36.32 & 31.16 & 34.89 & 5.27 & 5.86 & 62.25 & 57.95 & 54.73 & 7.16 & 11.06 & 42.64 & 41.62 & 41.18 & 8.55 & 9.32 \\
&                       & TP       & 7.70 & 6.99 & 7.50 & 1.70 & 0.55 & 8.16 & 8.26 & 7.63 & 0.59 & 1.46 & 3.74 & 3.47 & 3.57 & 1.17 & 2.75 \\
&                       & MEND     & 31.84 & 26.60 & 33.98 & 43.08 & 44.90 & 49.30 & 51.46 & 46.70 & \underline{81.03} & 82.53 & 42.35 & 45.07 & 42.68 & \underline{66.98} & 62.32 \\
&                       & SERAC    & 84.50 & 60.36 & 81.83 & \underline{63.26} & 13.40 & 47.03 & 45.79 & 44.82 & 42.12 & 15.31 & 57.02 & 51.66 & 56.80 & 48.48 & 13.24 \\
&                       & LiveEdit & \underline{92.75} & \textbf{89.81} & \underline{85.12} & 100.00 & \underline{96.82} & \underline{81.43} & \underline{80.34} & \textbf{62.25} & 100.00 & \underline{99.73} & \underline{93.41} & \textbf{89.87} & \underline{85.21} & 100.00 & \textbf{100.00} \\
&                       & \vicecell{VICE} & \vicecell{\textbf{99.98}} & \vicecell{\underline{88.37}} & \vicecell{\textbf{96.82}} & \vicecell{\textbf{100.00}} & \vicecell{\textbf{98.26}} & \vicecell{\textbf{98.93}} & \vicecell{\textbf{93.67}} & \vicecell{\underline{57.31}} & \vicecell{\textbf{100.00}} & \vicecell{\textbf{99.92}} & \vicecell{\textbf{95.27}} & \vicecell{\underline{82.95}} & \vicecell{\textbf{92.14}} & \vicecell{\textbf{100.00}} & \vicecell{\underline{98.29}} \\
\bottomrule
\end{tabular}%
}
\caption{Main results of lifelong editing for LLaVA-v1.5 and MiniGPT4 on EVQA, EIC, and VLKEB. ``\#Edits'' represents the number of new facts. MEND, SERAC, and LiveEdit results are from \citet{chen2025liveedit}; FT-VL, TP, and \vice{} are from our runs.}
\label{tab:main_results}
\end{table*}

\noindent\textbf{Learnable scope-discriminative head.} The scope-discriminative head $h_{\psi}$ is a two-layer MLP and is trained offline,
separately from the editing and inference pipeline, while both CLIP and the MLLM remain frozen.
Training uses no test edit anchors or evaluation queries.
For each training edit group, the reliability entry is used as the anchor, text/image-rephrase samples are treated as in-scope examples, and text/image-locality samples are treated as out-of-scope examples.
Given minibatch anchors $\mathcal{A}$, non-anchor in-scope examples $\mathcal{R}$, and locality examples $\mathcal{O}$, we optimize an anchor retrieval loss:
\begin{equation}
\mathcal{L}_{\text{ret}} = \frac{1}{|\mathcal{R}|} \sum_{q \in \mathcal{R}} -\log \frac{\exp(\mathbf{z}_q^{\top}\mathbf{a}^*(q) / T)}{\sum_{j \in \mathcal{A}} \exp(\mathbf{z}_q^{\top} \mathbf{a}_j / T)},
\end{equation}
where $\mathbf{a}^*(q)$ is the ground-truth anchor for $q$ and $T$ is the contrastive temperature.
We also use a margin loss to push locality examples away from all anchors:
\begin{equation}
\mathcal{L}_{\text{margin}} = \frac{1}{|\mathcal{O}|} \sum_{q \in \mathcal{O}} \max\bigl(0, \max_{j \in \mathcal{A}} \mathbf{z}_q^{\top} \mathbf{a}_j - m\bigr).
\end{equation}
The total objective is $\mathcal{L}_{\text{scope}} = \mathcal{L}_{\text{ret}} + \mathcal{L}_{\text{margin}}$.
After training, the dataset-specific $h_{\psi}$ is fixed and used only for retrieval and threshold routing.
Training details are provided in Appendix \ref{sec:appendix_vice_details}.

\section{Experiments}
\label{sec:experiments}

We evaluate \vice{} across three multimodal knowledge editing benchmarks and two vision-language backbones, comparing against five representative editing baselines. \Cref{sec:exp_setup} describes the experimental setup, including datasets, backbones, baselines, evaluation metrics, and implementation details. \Cref{sec:main_results} presents the main evaluation results, and \Cref{sec:ablations} reports our ablation studies.

\subsection{Experimental Setup}
\label{sec:exp_setup}

\paragraph{Datasets and Backbones.}
We evaluate on three standard multimodal knowledge editing benchmarks:
\textbf{EVQA}~\citep{cheng2023mmedit}, \textbf{EIC}~\citep{cheng2023mmedit}, and \textbf{VLKEB}~\citep{huang2024vlkeb}. We use two MLLMs as the backbone:
\textbf{LLaVA-v1.5-7B}~\citep{liu2024llava-v1.5} and \textbf{MiniGPT-4-vicuna-7B}~\citep{zhu2024minigpt-4}.
Both are widely used in the multimodal knowledge editing literature and represent different architectural choices.

\paragraph{Baselines.}
We compare \vice{} against five representative knowledge editing methods: \textbf{FT-VL}, fine-tuning the final layer of the MLLM; \textbf{TP}~\citep{huang2023transformer-patcher}, appending transformer-patchers to FFNs; \textbf{MEND}~\citep{mitchell2022mend}, training a hyper-network to learn to update model parameters; \textbf{SERAC}~\citep{mitchell2022serac}, a representative memory-based method; \textbf{LiveEdit}~\citep{chen2025liveedit}, a recent multimodal editing baseline that modulates hidden-states with MoE modules.

\paragraph{Metrics.}
Following~\citet{cheng2023mmedit}, we report five standard metrics:
\textbf{Reliability}: whether the model produces the target answer on the edited input;
\textbf{Text/Image-Locality}: whether unrelated text and multimodal queries remain unchanged;
\textbf{Text/Image-Generality}: whether the edit transfers to textually rephrased and visually similar variants.

\paragraph{Implementation details.}
For \vice{}, all MLLM backbone parameters are frozen throughout editing and inference.
We use CLIP-ViT-B/32 as $\phi$ and train only the two-layer head $h_\psi$ with $256$-dimensional outputs.
The head is trained offline for each dataset with AdamW for $80$ epochs, batch size $32$, learning rate $10^{-3}$, temperature $T=0.07$, and margin $m=0.5$.
Image summaries are generated before editing and cached.
We use the fixed threshold $\theta=0.65$ for all main experiments.

\begin{table}[t]
\centering
\footnotesize
\setlength{\tabcolsep}{3.2pt}
\resizebox{\columnwidth}{!}{%
\begin{tabular}{llccccc}
\toprule
\textbf{\# Edits} & \textbf{Ablation} & Rel & T-Gen & I-Gen & T-Loc & I-Loc \\
\midrule
\multirow{4}{*}{1}
& Full \vice{} & \textbf{1.00} & \textbf{0.88} & \textbf{0.97} & \textbf{1.00} & \textbf{0.99} \\
& w/o Scope Head & 1.00 & 0.86 & 0.96 & 1.00 & 0.85 \\
& w/o Image Summary & 0.75 & 0.46 & 0.80 & 1.00 & 1.00 \\
& w/o Retrieved Demos & 0.32 & 0.30 & 0.27 & 1.00 & 1.00 \\
\midrule
\multirow{4}{*}{1000}
& Full \vice{} & \textbf{1.00} & \textbf{0.88} & \textbf{0.97} & \textbf{1.00} & \textbf{0.98} \\
& w/o Scope Head & 1.00 & 0.85 & 0.95 & 1.00 & 0.76 \\
& w/o Image Summary & 0.75 & 0.46 & 0.80 & 0.99 & 1.00 \\
& w/o Retrieved Demos & 0.32 & 0.30 & 0.27 & 1.00 & 1.00 \\
\bottomrule
\end{tabular}
}
\caption{Ablation study of \vice{} on EVQA with LLaVA-v1.5. ``w/o Scope Head'' uses the raw retrieval route without the learned head $h_\psi$, ``w/o Image Summary'' removes the generated visual evidence summaries from demonstrations, and ``w/o Retrieved Demos'' removes retrieved visual in-context demonstrations.}
\label{tab:ablations}
\end{table}

\subsection{Main Results}
\label{sec:main_results}

As demonstrated in \Cref{tab:main_results}, \vice{} achieves strong and stable performance across diverse lifelong editing setups.
Its main gains appear in reliability, image-generality, and locality, which are central to visually grounded editing.
For example, under $1000$ edits for \llava{}-v1.5, \vice{} improves I-Gen. over \liveedit{} by $12.7$ on \evqa{}, $3.9$ on \eic{},
and $14.0$ on \vlkeb{}, and for \minigpt{}, by $11.7$ on \evqa{} and $6.9$ on \vlkeb{}.
\vice{} is not uniformly best on text-generality, where \liveedit{} remains competitive in some settings, but it maintains strong T-Gen., while giving more consistent image-side generalization and locality.
More importantly, \vice{} remains highly stable as the number of edits increases,
showing little performance degradation across both backbones and benchmarks.
By contrast, the baselines reveal a clear scalability--locality trade-off.
FT-VL often fits isolated edits but quickly harms locality under long sequences, while TP has limited capacity for complex multimodal knowledge.
MEND and SERAC are stronger in some single-edit settings, but MEND's accumulated updates scale poorly and SERAC's text-based retrieval gives weak image locality.
LiveEdit is the strongest recent baseline in multimodal knowledge editing, yet it still declines as edits accumulate, whereas \vice{} remains more stable and therefore better suited for lifelong multimodal knowledge editing.

\begin{figure}[t]
\centering
\includegraphics[width=\linewidth]{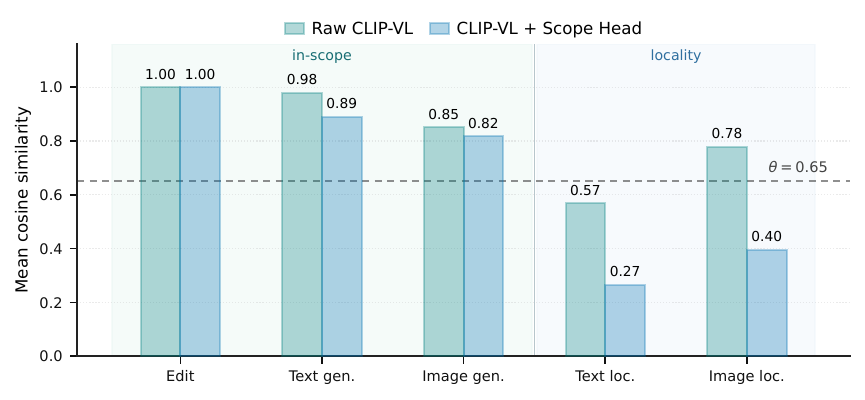}
\caption{Illustration of mean retrieval similarity of raw CLIP-VL features and learned scope-head embeddings on EVQA. For edit, text-rephrase, and image-rephrase queries, we report similarity to the paired edit anchor; for locality, we report nearest-anchor similarity.}
\label{fig:scope_similarity}
\end{figure}

\subsection{Robustness on \wbedit{}}

The diagnosis in \Cref{sec:diagnosis} shows that \ft{} and \liveedit{} can achieve reasonable average I-Gen.
while still varying strongly across target-context groups. Applying the same analysis to \vice{},
\Cref{tab:p4_results} shows a different pattern: \vice{} reaches the highest overall I-Gen. ($93.4$) and the smallest PG ($4.7$).
Its group accuracies also remain consistently high,
rather than showing a sharp worst-group drop.
This suggests that \vice{} improves not only average I-Gen., but also worst-group robustness, reducing sensitivity to background context.
The observation on \wbedit{} consists with the main results on \evqa{}/\eic{/\vlkeb{}}: methods with lower PG under background shifts on \wbedit{} generally achieve stronger I-Gen. on standard benchmarks.

Object-only masking provides a complementary check. When the background is removed and the bird region is preserved,
\vice{} remains strong ($89.0$) and does not show the large masking benefit observed for the baselines.
This indicates that its original image-rephrase performance is less degraded by spurious context.
Together, the group-level and masking results support the claim that visual in-context editing reduces evidence misgrounding on \wbedit{}.

\subsection{Ablation Study}
\label{sec:ablations}

To assess the contribution of each component, we ablate the scope head, image summaries, and retrieved demonstrations on EVQA with LLaVA-v1.5.
As shown in \Cref{tab:ablations}, removing the scope head mainly reduces I-Loc.,
which suggests that scope routing helps prevent irrelevant edit application.
Please refer to Appendix~\ref{sec:appendix_additional_discussion} for a detailed discussion of retrieval and routing quality.
Removing image summaries lowers Rel., T-Gen., and I-Gen.,
showing that explicit visual evidence is important for generalizing the edit to rephrased or visually similar inputs.
We also provide a discussion of the summary quality in Appendix~\ref{sec:appendix_additional_discussion}.
The largest drop occurs when retrieved demonstrations are removed,
indicating that they provide the main in-context editing signal.
Overall, the ablation results align with the intended roles of the three components:
retrieved demonstrations convey the edit, image summaries ground it in visual evidence, and the scope-discriminative head preserves locality.

\Cref{fig:scope_similarity} further supports this interpretation:
the learned scope head creates a clearer margin between in-scope and out-of-scope facts,
whereas raw CLIP-VL assigns some locality queries similarities close to in-scope examples, making them more likely to trigger irrelevant edits.

\subsection{Sensitivity to Similarity Threshold}
\label{sec:threshold}

\begin{figure}[t]
\centering
\includegraphics[width=0.85\linewidth]{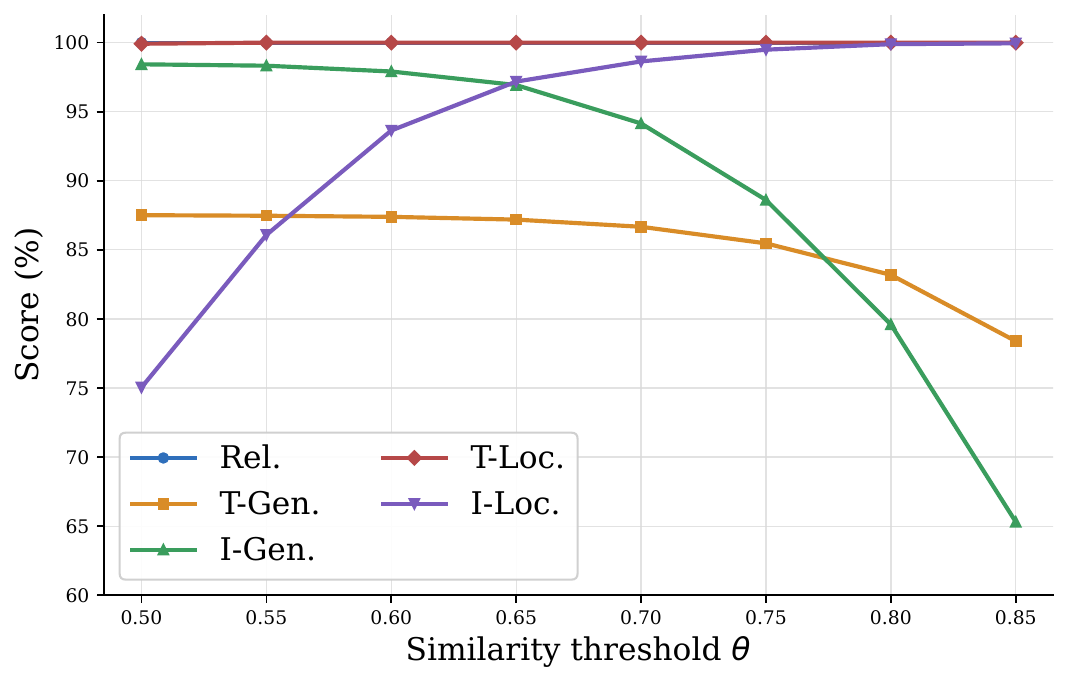}
\caption{Five standard metrics under different similarity thresholds $\theta$ on EVQA with LLaVA-v1.5.}
\label{fig:threshold_sweep}
\end{figure}

We evaluate the sensitivity of \vice{} to the similarity threshold $\theta$ by sweeping $\theta$ from $0.50$ to $0.85$ on EVQA with LLaVA-v1.5. As shown in \Cref{fig:threshold_sweep}, reliability and text-locality are stable across the sweep, while image-generality and image-locality show the main trade-off. A low threshold applies edits too broadly, giving high I-Gen. ($98.4$ at $\theta=0.50$) but low I-Loc. ($75.0$). A very high threshold is too conservative, causing I-Gen. to drop to $65.3$ at $\theta=0.85$. We use $\theta=0.65$ in the main experiments because it gives a good balance, with I-Gen. $96.9$ and I-Loc. $97.2$. We also keep the same $\theta=0.65$ for other main experiments to avoid additional hyperparameter tuning.

\section{Related Work}
\label{sec:related}

\paragraph{Knowledge Editing.}
Knowledge editing aims to update specific factual associations in pretrained models while preserving their behavior on unrelated inputs.
Prior work edits models by rewriting located parameters~\citep{dai2022knowledge-neurons,meng2022rome,meng2022memit,fang2025alphaedit}, learning parameter updates~\citep{sinitsin2020editable-neural-networks,de2021knowledge-editor,mitchell2022mend,tan2024malmen}, adding trainable modules~\citep{huang2023transformer-patcher,wang2024lemoe}, or using external memory and routing~\citep{mitchell2022serac,hartvigsen2023grace,yu2024melo,wang2024wise,zeng2025mscke,xia2025medrek}.
In-context editing~\citep{si2022prompting-gpt-3-to-be-reliable,madaan2022memory-prompt,zheng2023ike,ma2025comprehendedit} instead specifies new facts through demonstrations at inference time, avoiding backbone updates. To achieve lifelong editing~\citep{chen2025liveedit,chen2024lifelong,pan2024unike}, various recent works have incorporated lifelong learning techniques~\citep{wang2024cl_survey,wang2026singular,wu2025sd-lora,wu2024meta-cl}.

Multimodal knowledge editing~\citep{chen2025liveedit,shi2025dualedit} extends these ideas to vision-language models. MMEdit~\citep{cheng2023mmedit} and VLKEB~\citep{huang2024vlkeb} establish general evaluation protocols, while MIKE~\citep{li2024mikebenchmark} studies fine-grained multimodal entity editing across recognition and generation tasks. MC-MKE~\citep{zhang2025mcmke} decomposes multimodal knowledge into visual and textual components and evaluates consistency across modalities. FGVEdit~\citep{zhang2025mcmke} focuses on the fine-grained editing scope of particular entities in complex images. 
These evaluations mainly test answer correctness, not whether the edited answer is grounded in the intended visual evidence.
Our diagnostics are complementary: it keeps the desired fact fixed while systematically varying incidental background context to measure whether an answer-level successful edit remains context-sensitive.
Our method \vice{} addresses this gap with visual in-context demonstrations and a lightweight scope router.

\paragraph{In-Context Learning.}
In-context learning~\citep{brown2020language-models-are-few-shot-learners,liu2022what-makes-good-in-context-examples-for-gpt-3,bu2024provably,bu2025provable} enables large language models to adapt to new tasks from demonstrations without parameter updates.
Vision-language models can also learn from interleaved image-text demonstrations, though cross-modal interaction makes visual in-context learning more challenging~\citep{schwettmann2023multimodal-neurons-in-pretrained-text-only-transformers}.
Recent VICL work retrieves visual demonstrations and summarizes images according to task intent~\citep{zhou2024vicl}.
We adapt this idea to multimodal knowledge editing: \vice{} uses visual demonstrations to express edited facts and a scope head to decide when they should apply.

\section{Conclusion}
\label{sec:conclusion}

We studied multimodal knowledge editing beyond answer-level success, asking whether edited predictions follow the intended visual evidence. Standard metrics can hide evidence misgrounding: a model may produce the target answer while remaining sensitive to incidental context. To make this visible, we introduced \wbedit{}, which separates bird-habitat facts from background cues, and \vice{}, a frozen-backbone framework that stores edits in sequence memory and applies them through visual in-context demonstrations with scope routing. Across lifelong editing benchmarks, \vice{} remains strong; on \wbedit{}, it substantially reduces sensitivity to spurious background context. These results show that explicitly exposing target visual evidence helps make multimodal edits more grounded and scalable.

\subsection*{Limitations}

\wbedit{} is designed as a controlled diagnostic setting for bird-habitat editing, where target evidence and background context can be explicitly separated. This makes the evidence-misgrounding pattern measurable, while future work can extend the same diagnostic principle to broader visual facts and more open-ended scenes. \vice{} currently uses generated image summaries and a learned retrieval-based scope decision to expose and apply edits. These components are effective in our experiments, and further improvements in summary quality, retrieval calibration, and uncertainty handling may make the framework more reliable in ambiguous boundary cases. Finally, \vice{} shifts part of the editing burden from backbone updates to external memory, retrieval, and longer prompts. This trade-off preserves the frozen MLLM and supports lifelong editing, while leaving useful opportunities for memory compression, faster indexing, and more compact prompt construction.

\section*{Acknowledgments}

We would like to thank the anonymous reviewers for their insightful comments and constructive suggestions.
This work was supported by the Fundamental and
Interdisciplinary Disciplines Breakthrough Plan of the Ministry of Education of China (No.~JYB2025XDXM101), the National
Natural Science Foundation of China (No.~62306233),
the Tianyuan Fund for Mathematics of the National Natural
Science Foundation of China (No.~12426105), and the
National Natural Science Foundation of China
(No.~62476214, No.~12471485, No~62331028).

\bibliography{references}

@inproceedings{li2023blip-2,
  title={Blip-2: Bootstrapping language-image pre-training with frozen image encoders and large language models},
  author={Li, Junnan and Li, Dongxu and Savarese, Silvio and Hoi, Steven},
  booktitle={International Conference on Machine Learning},
  pages={19730--19742},
  year={2023},
}

@inproceedings{zhu2024minigpt-4,
  title={Minigpt-4: Enhancing vision-language understanding with advanced large language models},
  author={Zhu, Deyao and Shen, Xiaoqian and Li, Xiang and Elhoseiny, Mohamed and others},
  booktitle={International Conference on Learning Representations},
  volume={2024},
  pages={18378--18394},
  year={2024}
}

@article{liu2023llava,
  title={Visual instruction tuning},
  author={Liu, Haotian and Li, Chunyuan and Wu, Qingyang and Lee, Yong Jae},
  journal={Advances in neural information processing systems},
  volume={36},
  pages={34892--34916},
  year={2023}
}

@inproceedings{liu2024llava-v1.5,
  title={Improved baselines with visual instruction tuning},
  author={Liu, Haotian and Li, Chunyuan and Li, Yuheng and Lee, Yong Jae},
  booktitle={Proceedings of the IEEE/CVF conference on computer vision and pattern recognition},
  pages={26296--26306},
  year={2024}
}

@article{Qwen2-VL,
  title={Qwen2-VL: Enhancing Vision-Language Model's Perception of the World at Any Resolution},
  author={Wang, Peng and Bai, Shuai and Tan, Sinan and Wang, Shijie and Fan, Zhihao and Bai, Jinze and Chen, Keqin and Liu, Xuejing and Wang, Jialin and Ge, Wenbin and Fan, Yang and Dang, Kai and Du, Mengfei and Ren, Xuancheng and Men, Rui and Liu, Dayiheng and Zhou, Chang and Zhou, Jingren and Lin, Junyang},
  journal={arXiv preprint arXiv:2409.12191},
  year={2024}
}

@article{Qwen3-VL,
      title={Qwen3-VL Technical Report}, 
      author={Shuai Bai and Yuxuan Cai and Ruizhe Chen and Keqin Chen and Xionghui Chen and Zesen Cheng and Lianghao Deng and Wei Ding and Chang Gao and Chunjiang Ge and Wenbin Ge and Zhifang Guo and Qidong Huang and Jie Huang and Fei Huang and Binyuan Hui and Shutong Jiang and Zhaohai Li and Mingsheng Li and Mei Li and Kaixin Li and Zicheng Lin and Junyang Lin and Xuejing Liu and Jiawei Liu and Chenglong Liu and Yang Liu and Dayiheng Liu and Shixuan Liu and Dunjie Lu and Ruilin Luo and Chenxu Lv and Rui Men and Lingchen Meng and Xuancheng Ren and Xingzhang Ren and Sibo Song and Yuchong Sun and Jun Tang and Jianhong Tu and Jianqiang Wan and Peng Wang and Pengfei Wang and Qiuyue Wang and Yuxuan Wang and Tianbao Xie and Yiheng Xu and Haiyang Xu and Jin Xu and Zhibo Yang and Mingkun Yang and Jianxin Yang and An Yang and Bowen Yu and Fei Zhang and Hang Zhang and Xi Zhang and Bo Zheng and Humen Zhong and Jingren Zhou and Fan Zhou and Jing Zhou and Yuanzhi Zhu and Ke Zhu},
	  journal={arXiv preprint arXiv:2511.21631},
      year={2025}
}

@article{wah2011cub,
  title={{The Caltech-Ucsd Birds-200-2011 Dataset}},
  author={Wah, Catherine and Branson, Steve and Welinder, Peter and Perona, Pietro and Belongie, Serge},
  year={2011},
  publisher={California Institute of Technology}
}

@article{kwiatkowski2019natural-questions,
  title={Natural questions: a benchmark for question answering research},
  author={Kwiatkowski, Tom and Palomaki, Jennimaria and Redfield, Olivia and Collins, Michael and Parikh, Ankur and Alberti, Chris and Epstein, Danielle and Polosukhin, Illia and Devlin, Jacob and Lee, Kenton and others},
  journal={Transactions of the Association for Computational Linguistics},
  volume={7},
  pages={453--466},
  year={2019},
  publisher={MIT Press One Rogers Street, Cambridge, MA 02142-1209, USA journals-info~…}
}

@inproceedings{marino2019ok-vqa,
  title={Ok-vqa: A visual question answering benchmark requiring external knowledge},
  author={Marino, Kenneth and Rastegari, Mohammad and Farhadi, Ali and Mottaghi, Roozbeh},
  booktitle={Proceedings of the IEEE/cvf conference on computer vision and pattern recognition},
  pages={3195--3204},
  year={2019}
}

@inproceedings{dai2022knowledge-neurons,
  title={Knowledge neurons in pretrained transformers},
  author={Dai, Damai and Dong, Li and Hao, Yaru and Sui, Zhifang and Chang, Baobao and Wei, Furu},
  booktitle={Proceedings of the 60th Annual Meeting of the Association for Computational Linguistics (Volume 1: Long Papers)},
  pages={8493--8502},
  year={2022}
}

@article{meng2022rome,
  title={Locating and editing factual associations in gpt},
  author={Meng, Kevin and Bau, David and Andonian, Alex and Belinkov, Yonatan},
  journal={Advances in neural information processing systems},
  volume={35},
  pages={17359--17372},
  year={2022}
}

@article{meng2022memit,
  title={Mass-editing memory in a transformer},
  author={Meng, Kevin and Sharma, Arnab Sen and Andonian, Alex and Belinkov, Yonatan and Bau, David},
  journal={arXiv preprint arXiv:2210.07229},
  year={2022}
}

@inproceedings{fang2025alphaedit,
  title={Alphaedit: Null-space constrained knowledge editing for language models},
  author={Fang, Junfeng and Jiang, Houcheng and Wang, Kun and Ma, Yunshan and Shi, Jie and Wang, Xiang and He, Xiangnan and Chua, Tat-Seng},
  booktitle={International Conference on Learning Representations},
  volume={2025},
  pages={16366--16396},
  year={2025}
}

@article{sinitsin2020editable-neural-networks,
  title={Editable neural networks},
  author={Sinitsin, Anton and Plokhotnyuk, Vsevolod and Pyrkin, Dmitriy and Popov, Sergei and Babenko, Artem},
  journal={arXiv preprint arXiv:2004.00345},
  year={2020}
}

@inproceedings{de2021knowledge-editor,
  title={Editing factual knowledge in language models},
  author={De Cao, Nicola and Aziz, Wilker and Titov, Ivan},
  booktitle={Proceedings of the 2021 conference on empirical methods in natural language processing},
  pages={6491--6506},
  year={2021}
}

@inproceedings{mitchell2022mend,
  title={Fast Model Editing at Scale},
  author={Mitchell, Eric and Lin, Charles and Bosselut, Antoine and Finn, Chelsea and Manning, Christopher D},
  booktitle={International Conference on Learning Representations},
  year={2022},
  pages={11576--11597}
}

@inproceedings{tan2024malmen,
  title={Massive editing for large language models via meta learning},
  author={Tan, Chenmien and Zhang, Ge and Fu, Jie},
  booktitle={International Conference on Learning Representations},
  volume={2024},
  pages={54676--54690},
  year={2024}
}

@inproceedings{mitchell2022serac,
  title={Memory-based model editing at scale},
  author={Mitchell, Eric and Lin, Charles and Bosselut, Antoine and Manning, Christopher D and Finn, Chelsea},
  booktitle={International Conference on Machine Learning},
  pages={15817--15831},
  year={2022}
}

@article{hartvigsen2023grace,
  title={Aging with grace: Lifelong model editing with discrete key-value adaptors},
  author={Hartvigsen, Tom and Sankaranarayanan, Swami and Palangi, Hamid and Kim, Yoon and Ghassemi, Marzyeh},
  journal={Advances in Neural Information Processing Systems},
  volume={36},
  pages={47934--47959},
  year={2023}
}

@inproceedings{yu2024melo,
  title={Melo: Enhancing model editing with neuron-indexed dynamic lora},
  author={Yu, Lang and Chen, Qin and Zhou, Jie and He, Liang},
  booktitle={Proceedings of the AAAI Conference on Artificial Intelligence},
  volume={38},
  number={17},
  pages={19449--19457},
  year={2024}
}

@article{wang2024wise,
  title={Wise: Rethinking the knowledge memory for lifelong model editing of large language models},
  author={Wang, Peng and Li, Zexi and Zhang, Ningyu and Xu, Ziwen and Yao, Yunzhi and Jiang, Yong and Xie, Pengjun and Huang, Fei and Chen, Huajun},
  journal={Advances in Neural Information Processing Systems},
  volume={37},
  pages={53764--53797},
  year={2024}
}

@inproceedings{zeng2025mscke,
  title={Visual-oriented fine-grained knowledge editing for multimodal large language models},
  author={Zeng, Zhen and Gu, Leijiang and Yang, Xun and Duan, Zhangling and Shi, Zenglin and Wang, Meng},
  booktitle={Proceedings of the IEEE/CVF International Conference on Computer Vision},
  pages={2491--2500},
  year={2025}
}

@article{xia2025medrek,
  title={MedREK: Retrieval-Based Editing for Medical LLMs with Key-Aware Prompts},
  author={Xia, Shujun and Lin, Haokun and Wu, Yichen and Zhou, Yinan and Li, Zixuan and Wan, Zhongwei and Xing, Xingrun and Zheng, Yefeng and Li, Xiang and Shan, Caifeng and others},
  journal={arXiv preprint arXiv:2510.13500},
  year={2025}
}

@inproceedings{chen2024lifelong,
  title={Lifelong knowledge editing for llms with retrieval-augmented continuous prompt learning},
  author={Chen, Qizhou and Zhang, Taolin and He, Xiaofeng and Li, Dongyang and Wang, Chengyu and Huang, Longtao and others},
  booktitle={Proceedings of the 2024 Conference on Empirical Methods in Natural Language Processing},
  pages={13565--13580},
  year={2024}
}

@article{pan2024unike,
  title={Towards unified multimodal editing with enhanced knowledge collaboration},
  author={Pan, Kaihang and Fan, Zhaoyu and Li, Juncheng and Yu, Qifan and Fei, Hao and Tang, Siliang and Hong, Richang and Zhang, Hanwang and Sun, Qianru},
  journal={Advances in Neural Information Processing Systems},
  volume={37},
  pages={110290--110314},
  year={2024}
}

@article{si2022prompting-gpt-3-to-be-reliable,
  title={Prompting gpt-3 to be reliable},
  author={Si, Chenglei and Gan, Zhe and Yang, Zhengyuan and Wang, Shuohang and Wang, Jianfeng and Boyd-Graber, Jordan and Wang, Lijuan},
  journal={arXiv preprint arXiv:2210.09150},
  year={2022}
}

@inproceedings{madaan2022memory-prompt,
  title={Memory-assisted prompt editing to improve GPT-3 after deployment},
  author={Madaan, Aman and Tandon, Niket and Clark, Peter and Yang, Yiming},
  booktitle={Proceedings of the 2022 Conference on Empirical Methods in Natural Language Processing},
  pages={2833--2861},
  year={2022}
}

@inproceedings{zheng2023ike,
  title={Can we edit factual knowledge by in-context learning?},
  author={Zheng, Ce and Li, Lei and Dong, Qingxiu and Fan, Yuxuan and Wu, Zhiyong and Xu, Jingjing and Chang, Baobao},
  booktitle={Proceedings of the 2023 Conference on Empirical Methods in Natural Language Processing},
  pages={4862--4876},
  year={2023}
}

@inproceedings{ma2025comprehendedit,
  title={Comprehendedit: a comprehensive dataset and evaluation framework for multimodal knowledge editing},
  author={Ma, Yaohui and Hong, Xiaopeng and Zhang, Shizhou and Li, Huiyun and Zhu, Zhilin and Luo, Wei and Ma, Zhiheng},
  booktitle={Proceedings of the AAAI Conference on Artificial Intelligence},
  volume={39},
  number={18},
  pages={19323--19331},
  year={2025}
}

@inproceedings{huang2023transformer-patcher,
  title={Transformer-patcher: One mistake worth one neuron},
  author={Huang, Zeyu and Shen, Yikang and Zhang, Xiaofeng and Zhou, Jie and Rong, Wenge and Xiong, Zhang},
  booktitle={International Conference on Learning Representations},
  year={2023},
  pages={13927--13943}
}

@inproceedings{wang2024lemoe,
  title={Lemoe: Advanced mixture of experts adaptor for lifelong model editing of large language models},
  author={Wang, Renzhi and Li, Piji},
  booktitle={Proceedings of the 2024 Conference on Empirical Methods in Natural Language Processing},
  pages={2551--2575},
  year={2024}
}

@inproceedings{cheng2023mmedit,
  title={Can we edit multimodal large language models?},
  author={Cheng, Siyuan and Tian, Bozhong and Liu, Qingbin and Chen, Xi and Wang, Yongheng and Chen, Huajun and Zhang, Ningyu},
  booktitle={Proceedings of the 2023 conference on empirical methods in natural language processing},
  pages={13877--13888},
  year={2023}
}

@article{huang2024vlkeb,
  title={Vlkeb: A large vision-language model knowledge editing benchmark},
  author={Huang, Han and Zhong, Haitian and Yu, Tao and Liu, Qiang and Wu, Shu and Wang, Liang and Tan, Tieniu},
  journal={Advances in Neural Information Processing Systems},
  volume={37},
  pages={9257--9280},
  year={2024}
}

@inproceedings{chen2025liveedit,
  title={Lifelong knowledge editing for vision language models with low-rank mixture-of-experts},
  author={Chen, Qizhou and Wang, Chengyu and Wang, Dakan and Zhang, Taolin and Li, Wangyue and He, Xiaofeng},
  booktitle={Proceedings of the Computer Vision and Pattern Recognition Conference},
  pages={9455--9466},
  year={2025}
}

@article{shi2025dualedit,
  title={DualEdit: Dual Editing for Knowledge Updating in Vision-Language Models},
  author={Shi, Zhiyi and Wang, Binjie and Si, Chongjie and Wu, Yichen and Kim, Junsik and Pfister, Hanspeter},
  journal={arXiv preprint arXiv:2506.13638},
  year={2025}
}

@inproceedings{li2024mikebenchmark,
  title={Mike: A new benchmark for fine-grained multimodal entity knowledge editing},
  author={Li, Jiaqi and Du, Miaozeng and Zhang, Chuanyi and Chen, Yongrui and Hu, Nan and Qi, Guilin and Jiang, Haiyun and Cheng, Siyuan and Tian, Bozhong},
  booktitle={Findings of the Association for Computational Linguistics: ACL 2024},
  pages={5018--5029},
  year={2024}
}

@inproceedings{zhang2025mcmke,
  title={Mc-mke: A fine-grained multimodal knowledge editing benchmark emphasizing modality consistency},
  author={Zhang, Junzhe and Zhang, Huixuan and Yin, Xunjian and Huang, Baizhou and Zhang, Xu and Hu, Xinyu and Wan, Xiaojun},
  booktitle={Findings of the Association for Computational Linguistics: ACL 2025},
  pages={17430--17445},
  year={2025}
}

@article{brown2020language-models-are-few-shot-learners,
  title={Language models are few-shot learners},
  author={Brown, Tom and Mann, Benjamin and Ryder, Nick and Subbiah, Melanie and Kaplan, Jared D and Dhariwal, Prafulla and Neelakantan, Arvind and Shyam, Pranav and Sastry, Girish and Askell, Amanda and others},
  journal={Advances in neural information processing systems},
  volume={33},
  pages={1877--1901},
  year={2020}
}

@inproceedings{liu2022what-makes-good-in-context-examples-for-gpt-3,
  title={What makes good in-context examples for GPT-3?},
  author={Liu, Jiachang and Shen, Dinghan and Zhang, Yizhe and Dolan, William B and Carin, Lawrence and Chen, Weizhu},
  booktitle={Proceedings of Deep Learning Inside Out (DeeLIO 2022): The 3rd workshop on knowledge extraction and integration for deep learning architectures},
  pages={100--114},
  year={2022}
}

@inproceedings{schwettmann2023multimodal-neurons-in-pretrained-text-only-transformers,
  title={Multimodal neurons in pretrained text-only transformers},
  author={Schwettmann, Sarah and Chowdhury, Neil and Klein, Samuel and Bau, David and Torralba, Antonio},
  booktitle={Proceedings of the IEEE/CVF International Conference on Computer Vision},
  pages={2862--2867},
  year={2023}
}

@inproceedings{zhou2024vicl,
  title={Visual in-context learning for large vision-language models},
  author={Zhou, Yucheng and Li, Xiang and Wang, Qianning and Shen, Jianbing},
  booktitle={Findings of the Association for Computational Linguistics: ACL 2024},
  pages={15890--15902},
  year={2024}
}

@article{bu2024provably,
  title={Provably transformers harness multi-concept word semantics for efficient in-context learning},
  author={Bu, Dake and Huang, Wei and Han, Andi and Nitanda, Atsushi and Suzuki, Taiji and Zhang, Qingfu and Wong, Hau-San},
  journal={Advances in Neural Information Processing Systems},
  volume={37},
  pages={63342--63405},
  year={2024}
}

@article{bu2025provable,
  title={Provable in-context vector arithmetic via retrieving task concepts},
  author={Bu, Dake and Huang, Wei and Han, Andi and Nitanda, Atsushi and Zhang, Qingfu and Wong, Hau-San and Suzuki, Taiji},
  journal={arXiv preprint arXiv:2508.09820},
  year={2025}
}

@article{wang2024cl_survey,
  title={A comprehensive survey of continual learning: Theory, method and application},
  author={Wang, Liyuan and Zhang, Xingxing and Su, Hang and Zhu, Jun},
  journal={IEEE transactions on pattern analysis and machine intelligence},
  volume={46},
  number={8},
  pages={5362--5383},
  year={2024},
  publisher={IEEE}
}

@article{wang2026singular,
  title={Singular value fine-tuning for few-shot class-incremental learning},
  author={Wang, Zhiwu and Wu, Yichen and Wang, Renzhen and Lin, Haokun and Wang, Quanziang and Zhao, Qian and Meng, Deyu},
  journal={IEEE Transactions on Circuits and Systems for Video Technology},
  year={2026},
  publisher={IEEE}
}

@article{wu2025sd-lora,
  title={Sd-lora: Scalable decoupled low-rank adaptation for class incremental learning},
  author={Wu, Yichen and Piao, Hongming and Huang, Long-Kai and Wang, Renzhen and Li, Wanhua and Pfister, Hanspeter and Meng, Deyu and Ma, Kede and Wei, Ying},
  journal={arXiv preprint arXiv:2501.13198},
  year={2025}
}

@inproceedings{wu2024meta-cl,
  title={Meta continual learning revisited: Implicitly enhancing online hessian approximation via variance reduction},
  author={Wu, Yichen and Huang, Long-Kai and Wang, Renzhen and Meng, Deyu and Wei, Ying},
  booktitle={The Twelfth international conference on learning representations},
  year={2024}
}

@inproceedings{sagawa2020group-dro,
  title={Distributionally Robust Neural Networks},
  author={Sagawa, Shiori and Koh, Pang Wei and Hashimoto, Tatsunori B and Liang, Percy},
  booktitle={International Conference on Learning Representations},
  year={2020},
  pages={2661--2680}
}

\clearpage
\appendix

\section{\wbedit{} Construction Details}
\label{sec:appendix_wbedit_construction}

\paragraph{Source data and labels.}
\wbedit{} is constructed from \cub{}, which provides fine-grained bird images with species identities.
We convert the original species labels into a binary habitat label
$y\in\{\text{water},\text{land}\}$.
We obtain the background label $b\in\{\text{water},\text{land}\}$ by using a pretrained vision-language model to judge whether the surrounding scene contains visible water.
Specifically, we query LLaVA-v1.6-Vicuna-13B on each CUB image with a binary prompt asking whether the background contains any water body, such as a river, lake, ocean, pond, waterfall, or stream.
The model is instructed to answer only ``Yes'' or ``No''; we map ``Yes'' to a water background and ``No'' to a land background.
The generated file contains one background label for each of the $11{,}788$ CUB images.
These background labels are used only for dataset construction and group-wise diagnostics, and are not included in the edit prompt.
This gives an explicit target-context grouping $(y,b)$, where the first component is the bird habitat label and the second component is the surrounding background.
The construction is designed so that the edited answer should follow the bird identity rather than the background scene.

\paragraph{Background-labeling prompt.}
The key prompt used for assigning background labels is:
\begin{quote}
\small
\texttt{Please carefully examine the background of this image and determine whether there is any water present (such as rivers, lakes, oceans, ponds, waterfalls, streams, or any other form of water). Please respond with only "Yes" or "No", without any additional explanations or descriptions. If you can see any water body in the background, answer "Yes"; if there is no water body visible in the background, answer "No".}
\end{quote}

\paragraph{Edit-image filtering.}
We start from a filtered CUB pool that contains images on which the pretrained \llava{}-v1.5-7B and \minigpt-vicuna-7B fail the binary habitat classification task.
This filtering produces challenging edit requests where the target answer corrects a pre-edit failure, rather than merely confirming knowledge that the model already expresses reliably.

\paragraph{Evaluation data construction.}
Each retained image is converted into a standard multimodal editing instance with one edit request, two generality probes, and two locality probes.
The construction of each field is as follows.

\noindent\textit{Reliability.}
The edit request uses the retained failure image and the source prompt
\begin{quote}
\small
\texttt{What type of bird is in the image? Choose from: land, water.}
\end{quote}
with the corrected habitat label as the new target answer.
This field tests whether an editor can correct the pre-edit failure on the original edit input.

\noindent\textit{Text-generality.}
Text-generality keeps the same edit image and target answer, but replaces the source prompt with a paraphrased binary habitat question.
We randomly sample from the following paraphrase templates:
\begin{quote}
\small
\texttt{1. Based on its physical features, would you classify this bird as a water bird or a land bird?}

\texttt{2. Does this bird primarily inhabit aquatic environments or terrestrial ones? Answer with 'water' or 'land'.}

\texttt{3. Identify the habitat type of this bird: water or land?}

\texttt{4. Is the bird shown here adapted for life in water or on land?}
\end{quote}
These paraphrases ask for the same water-vs.-land habitat decision without mentioning the background label.
This field tests whether the edit transfers across textual reformulations of the same visual fact.

\noindent\textit{Image-generality.}
Image-generality keeps the source prompt and target answer, but replaces the edit image with a different image of the same bird species from the full CUB pool.
Because the species is unchanged, the desired habitat answer remains unchanged.
The replacement image may preserve or change the background type, which lets us test whether the edited answer follows the bird species rather than the surrounding scene.
This field is the main diagnostic probe: a robust edit should follow the bird species across image changes rather than switch according to the new background.

\noindent\textit{Text-locality.}
Following MMEdit~\citep{cheng2023mmedit}, text-locality uses unrelated question-answer pairs from \nq~\citep{kwiatkowski2019natural-questions}.
They test whether the edit changes unrelated text-side knowledge.

\noindent\textit{Image-locality.}
Following MMEdit~\citep{cheng2023mmedit}, image-locality uses unrelated multimodal question-answer examples from OK-VQA~\citep{marino2019ok-vqa}.
They test whether the edit changes unrelated multimodal behavior.

\paragraph{Splits and group distribution.}
The final \wbedit{} annotation file contains $1{,}136$ edit instances over $90$ bird species, split into $636$ training instances and $500$ test instances.
The main diagnostic results use the $500$ test instances as a sequential editing stream.
The test edit images remain strongly target-context imbalanced, which reflects the spurious correlations present in the filtered failure pool.
By contrast, the same-species image-rephrase set gives substantially broader coverage of the four target-context groups.
\Cref{tab:appendix_wbedit_groups} reports the group counts used in the test split.

\begin{table}[h]
\centering
\small
\setlength{\tabcolsep}{6pt}
\begin{tabular}{lrrrr}
\toprule
\textbf{Image Set} & \textbf{WW} & \textbf{WL} & \textbf{LW} & \textbf{LL} \\
\midrule
Edit images & 18 & 190 & 107 & 185 \\
Image-rephrase images & 115 & 93 & 133 & 159 \\
\bottomrule
\end{tabular}
\caption{\wbedit{} test-split group counts.
Each group is written as (target habitat, background), where W and L denote water and land.
The image-rephrase set is used for the group Image-Generality and PG diagnostics in \Cref{sec:diagnosis}.}
\label{tab:appendix_wbedit_groups}
\end{table}

\paragraph{CUB species-level habitat labels.}
\Cref{tab:appendix_cub_habitat_labels} lists the binary habitat label assigned to each of the 200 CUB species classes.
All images of the same species have the same $y$ label.

\section{Additional Implementation Details for \vice{}}
\label{sec:appendix_vice_details}

\paragraph{Image-summary generation prompt.}
Image summaries are generated offline and cached before editing.
Following the main text, we instantiate the frozen summary model $G$ as a pretrained MLLM before editing; in our preprocessing script, $G$ is implemented with \llava{}-v1.5-7B.
Given an image and its prompt, the summary generator uses the following prompt.
The method section writes this compactly as $G(i,x)$; in the implementation, the \texttt{\{text\}} field is instantiated with the prompt only, without the target answer.
\begin{quote}
\small
\ttfamily
\raggedright
USER: \textless image\textgreater\\
Given the image and the text prompt, write a concise visual summary for knowledge editing.
The summary must identify the target-related visual evidence, describe other salient objects or attributes that could act as distractors, and briefly contrast the target evidence against those distractors.
Do not answer the question directly. Do not simply copy the text prompt.\\
Return exactly three lines in the following format:\\
Target Evidence: ...\\
Distractors: ...\\
Contrast: ...\\
Text Prompt: \{text\}\\
ASSISTANT:
\end{quote}
Generation is deterministic with greedy decoding and a maximum of $128$ new tokens.
The returned summary is stored together with the corresponding image path and prompt, and is reused when constructing the evidence-summary demonstrations.

\paragraph{Structured template of the augmented prompt.}
For an accepted query, the augmented prompt consists of three parts: retrieved demonstrations, the retrieved edit, and the current query. Each part follows a structured template with three fields: ``Image Summary'', ``New Fact'', and ``Prompt''. We generate image summaries for the demonstration images and the retrieved edit image. For the current query, we instead use a generic textual placeholder that instructs the model to attend to the input real image. The complete augmented-prompt template is shown below:
\begin{quote}
\small
\ttfamily
\raggedright
...\\
{[Demonstration $c_i$]}\\
Image Summary: \{summary of a demonstration image\}\\
New Fact: \{demonstration edit prompt\} \{demonstration edit target\}\\
Prompt: \{demonstration query prompt\} \{demonstration query target/original answer\}\\
...

[Retrieved edit record]\\
Image Summary: \{summary of the retrieved edit image\}\\
New Fact: \{retrieved edit prompt\} \{retrieved edit target\}\\
Prompt: \{retrieved edit prompt\} \{retrieved edit target\}

[Current query]\\
Image Summary: Use the given real image as the visual evidence for this query.\\
New Fact: \{retrieved edit prompt\} \{retrieved edit target\}\\
Prompt: \{current query prompt\}
\end{quote}

\paragraph{Scope-head training data.}
The scope-discriminative head $h_{\psi}$ is trained offline for each dataset using only the training split.
For each training edit instance, we form one scope group.
The reliability example is used as the anchor, and its text-rephrase and image-rephrase examples are treated as in-scope queries with the same group label.
Text-locality and image-locality examples are treated as out-of-scope queries; each locality example is assigned its own unique label so it is not used as a positive example for any edit anchor.

\paragraph{Scope features and projection head.}
We encode each query $(i,x)$ with CLIP-ViT-B/32 (\texttt{openai/clip-vit-base-patch32}).
The text prompt is encoded with the CLIP text encoder.
If an image is available, it is encoded with the CLIP image encoder; for text-only locality examples, we use a zero image feature.
Both text and image features are $\ell_2$-normalized, concatenated, and kept fixed during scope-head training.
The learned head is a two-layer MLP,
\begin{equation}
h_{\psi}(\cdot)=\mathrm{Linear}_{d\rightarrow 256}(\mathrm{ReLU}(\mathrm{Linear}_{d\rightarrow d}(\cdot))),
\end{equation}
where $d$ is the concatenated CLIP feature dimension.
The projected features are $\ell_2$-normalized before computing retrieval similarities.

\paragraph{Optimization objective.}
Training minibatches are sampled at the edit-group level: each batch contains $32$ edit groups by default and includes all scope examples from those groups.
Within a minibatch, reliability examples provide the candidate anchors.
For non-anchor in-scope examples, we use the anchor retrieval loss in \Cref{sec:method}, which is a cross-entropy objective over the minibatch anchors.
For locality examples, we use the margin loss in \Cref{sec:method}, which penalizes cases where the maximum similarity to any minibatch anchor exceeds $m$.
The full objective is the same as in \Cref{sec:method}:
\begin{equation}
\mathcal{L}_{\mathrm{scope}}=\mathcal{L}_{\mathrm{ret}}+\mathcal{L}_{\mathrm{margin}}.
\end{equation}
Unless otherwise specified, we use AdamW for $80$ epochs with learning rate $10^{-3}$, temperature $T=0.07$, projection dimension $256$, and margin $m=0.5$.
After training, the dataset-specific head is frozen and used only for memory retrieval and threshold routing.

\section{Complete Experimental Results}
\label{sec:appendix_complete_results}

This section provides the full lifelong editing results that complement the condensed main table in \Cref{sec:experiments}.
In addition to the $1$, $100$, and $1000$ edit settings reported in the main text, we include the intermediate $10$-edit setting for both \llava{}-v1.5 and \minigpt{}.

\section{Additional Discussion}
\label{sec:appendix_additional_discussion}

\paragraph{Image-summary quality.}
To assess summary quality, we audited the 500 cached summaries used by VICE for the WATERBIRD-EDIT. The audit examines format consistency and the presence of recognizable target-object and scene/context cues, as shown in \Cref{tab:summary_quality}. Format consistency measures whether each summary follows the required structured format. Target matching checks for recognizable target-object cues, such as \textit{bird}, \textit{beak}, \textit{wing}, \textit{feathers}, \textit{plumage}, \textit{head}, \textit{tail}, \textit{perched}, \textit{standing} and so on. Distractor matching checks for scene or context cues, such as \textit{background}, \textit{water}, \textit{pond}, \textit{rocks}, \textit{leaf}, \textit{tree}, \textit{branch}, \textit{sky}, \textit{ground}, \textit{fence}, \textit{surface}, \textit{vegetation} and so on. The results show that the summaries consistently follow the required format and usually capture both target and contextual cues.

\begin{table}[t]
    \centering
    \resizebox{\linewidth}{!}{
    \begin{tabular}{cc}
    \toprule
       Format Consistency  &  500/500 = 100\%\\
       Target Matching Rate  &  476/500 = 95.2\%\\
       Distractor Matching Rate  &  431/500 = 86.2\%\\
    \bottomrule
    \end{tabular}
    }
    \caption{Summary quality audit.}
    \label{tab:summary_quality}
\end{table}

\paragraph{Retrieval and routing quality.}
We separately evaluate the retrieval and routing stages on EVQA with LLaVA-v1.5 at $\theta=0.65$. For in-scope queries (Rel., T-Gen., and I-Gen.), top-1 retrieval accuracy measures whether the intended edit is ranked first, while routing accuracy measures whether the retrieved edit is accepted. Routing errors on these queries are false rejections. For locality queries, no intended edit should be applied, so top-1 retrieval accuracy is not applicable; routing accuracy instead measures whether the query is correctly rejected, and the corresponding errors are false acceptances. As shown in \Cref{tab:retrieval_routing_quality}, the intended edit is retrieved for at least 97.3\% of all in-scope queries, and routing accuracy remains above 96.9\%. Reliability and text-locality queries are routed without errors. Image-locality queries are the most challenging, with a 5.7\% false-acceptance rate, but most unrelated image queries are still correctly rejected. These results indicate that the scope head generally applies the intended edit to in-scope queries while preventing irrelevant edit application.

\begin{table}[t]
    \centering
    \resizebox{\linewidth}{!}{%
    \begin{tabular}{cccc}
    \toprule
      Query type  &  Retrieval top-1  &  Routing acc  &  Routing error  \\
    \hline
      Rel.  &  100.0\%  &  100.0\%  &  false reject 0.0\%  \\
      T-Gen.  &  98.5\%  &  98.0\%  &  false reject 2.0\%  \\
      I-Gen.  &  97.3\%  &  96.9\%  &  false reject 3.1\%  \\
      T-Loc.  &  --     &  100.0\%  &  false accept 0.0\%  \\
      I-Loc.  &  --     &  94.3\%  &  false accept 5.7\%  \\
    \bottomrule
    \end{tabular}
    }
    \caption{VICE's Top-1 retrieval and routing accuracy on EVQA with LLaVA-v1.5 at $\theta=0.65$.}
    \label{tab:retrieval_routing_quality}
\end{table}

\onecolumn

\scriptsize
\setlength{\LTleft}{\fill}
\setlength{\LTright}{\fill}
\setlength{\tabcolsep}{1.5pt}
\renewcommand{\arraystretch}{0.95}
\begin{longtable}{@{}p{0.23\textwidth}p{0.23\textwidth}p{0.23\textwidth}p{0.23\textwidth}@{}}
\caption{CUB species-level habitat labels used by \wbedit{}.
}
\label{tab:appendix_cub_habitat_labels}\\
\toprule
\textbf{Class} & \textbf{Class} & \textbf{Class} & \textbf{Class} \\
\midrule
\endfirsthead
\toprule
\textbf{Class} & \textbf{Class} & \textbf{Class} & \textbf{Class} \\
\midrule
\endhead
\midrule
\multicolumn{4}{r}{Continued on next page} \\
\endfoot
\bottomrule
\endlastfoot
\textbf{001} Black footed Albatross (water) & \textbf{002} Laysan Albatross (water) & \textbf{003} Sooty Albatross (water) & \textbf{004} Groove billed Ani (land) \\
\textbf{005} Crested Auklet (water) & \textbf{006} Least Auklet (water) & \textbf{007} Parakeet Auklet (water) & \textbf{008} Rhinoceros Auklet (water) \\
\textbf{009} Brewer Blackbird (land) & \textbf{010} Red winged Blackbird (land) & \textbf{011} Rusty Blackbird (land) & \textbf{012} Yellow headed Blackbird (land) \\
\textbf{013} Bobolink (land) & \textbf{014} Indigo Bunting (land) & \textbf{015} Lazuli Bunting (land) & \textbf{016} Painted Bunting (land) \\
\textbf{017} Cardinal (land) & \textbf{018} Spotted Catbird (land) & \textbf{019} Gray Catbird (land) & \textbf{020} Yellow breasted Chat (land) \\
\textbf{021} Eastern Towhee (water) & \textbf{022} Chuck will Widow (land) & \textbf{023} Brandt Cormorant (water) & \textbf{024} Red faced Cormorant (water) \\
\textbf{025} Pelagic Cormorant (water) & \textbf{026} Bronzed Cowbird (land) & \textbf{027} Shiny Cowbird (land) & \textbf{028} Brown Creeper (land) \\
\textbf{029} American Crow (land) & \textbf{030} Fish Crow (land) & \textbf{031} Black billed Cuckoo (land) & \textbf{032} Mangrove Cuckoo (land) \\
\textbf{033} Yellow billed Cuckoo (land) & \textbf{034} Gray crowned Rosy Finch (land) & \textbf{035} Purple Finch (land) & \textbf{036} Northern Flicker (land) \\
\textbf{037} Acadian Flycatcher (land) & \textbf{038} Great Crested Flycatcher (land) & \textbf{039} Least Flycatcher (land) & \textbf{040} Olive sided Flycatcher (land) \\
\textbf{041} Scissor tailed Flycatcher (land) & \textbf{042} Vermilion Flycatcher (land) & \textbf{043} Yellow bellied Flycatcher (land) & \textbf{044} Frigatebird (water) \\
\textbf{045} Northern Fulmar (water) & \textbf{046} Gadwall (water) & \textbf{047} American Goldfinch (land) & \textbf{048} European Goldfinch (land) \\
\textbf{049} Boat tailed Grackle (land) & \textbf{050} Eared Grebe (water) & \textbf{051} Horned Grebe (water) & \textbf{052} Pied billed Grebe (water) \\
\textbf{053} Western Grebe (water) & \textbf{054} Blue Grosbeak (land) & \textbf{055} Evening Grosbeak (land) & \textbf{056} Pine Grosbeak (land) \\
\textbf{057} Rose breasted Grosbeak (land) & \textbf{058} Pigeon Guillemot (water) & \textbf{059} California Gull (water) & \textbf{060} Glaucous winged Gull (water) \\
\textbf{061} Heermann Gull (water) & \textbf{062} Herring Gull (water) & \textbf{063} Ivory Gull (water) & \textbf{064} Ring billed Gull (water) \\
\textbf{065} Slaty backed Gull (water) & \textbf{066} Western Gull (water) & \textbf{067} Anna Hummingbird (land) & \textbf{068} Ruby throated Hummingbird (land) \\
\textbf{069} Rufous Hummingbird (land) & \textbf{070} Green Violetear (land) & \textbf{071} Long tailed Jaeger (water) & \textbf{072} Pomarine Jaeger (water) \\
\textbf{073} Blue Jay (land) & \textbf{074} Florida Jay (land) & \textbf{075} Green Jay (land) & \textbf{076} Dark eyed Junco (land) \\
\textbf{077} Tropical Kingbird (land) & \textbf{078} Gray Kingbird (land) & \textbf{079} Belted Kingfisher (land) & \textbf{080} Green Kingfisher (land) \\
\textbf{081} Pied Kingfisher (land) & \textbf{082} Ringed Kingfisher (land) & \textbf{083} White breasted Kingfisher (land) & \textbf{084} Red legged Kittiwake (water) \\
\textbf{085} Horned Lark (land) & \textbf{086} Pacific Loon (water) & \textbf{087} Mallard (water) & \textbf{088} Western Meadowlark (water) \\
\textbf{089} Hooded Merganser (water) & \textbf{090} Red breasted Merganser (water) & \textbf{091} Mockingbird (land) & \textbf{092} Nighthawk (land) \\
\textbf{093} Clark Nutcracker (land) & \textbf{094} White breasted Nuthatch (land) & \textbf{095} Baltimore Oriole (land) & \textbf{096} Hooded Oriole (land) \\
\textbf{097} Orchard Oriole (land) & \textbf{098} Scott Oriole (land) & \textbf{099} Ovenbird (land) & \textbf{100} Brown Pelican (water) \\
\textbf{101} White Pelican (water) & \textbf{102} Western Wood Pewee (water) & \textbf{103} Sayornis (land) & \textbf{104} American Pipit (land) \\
\textbf{105} Whip poor Will (land) & \textbf{106} Horned Puffin (water) & \textbf{107} Common Raven (land) & \textbf{108} White necked Raven (land) \\
\textbf{109} American Redstart (land) & \textbf{110} Geococcyx (land) & \textbf{111} Loggerhead Shrike (land) & \textbf{112} Great Grey Shrike (land) \\
\textbf{113} Baird Sparrow (land) & \textbf{114} Black throated Sparrow (land) & \textbf{115} Brewer Sparrow (land) & \textbf{116} Chipping Sparrow (land) \\
\textbf{117} Clay colored Sparrow (land) & \textbf{118} House Sparrow (land) & \textbf{119} Field Sparrow (land) & \textbf{120} Fox Sparrow (land) \\
\textbf{121} Grasshopper Sparrow (land) & \textbf{122} Harris Sparrow (land) & \textbf{123} Henslow Sparrow (land) & \textbf{124} Le Conte Sparrow (land) \\
\textbf{125} Lincoln Sparrow (land) & \textbf{126} Nelson Sharp tailed Sparrow (land) & \textbf{127} Savannah Sparrow (land) & \textbf{128} Seaside Sparrow (land) \\
\textbf{129} Song Sparrow (land) & \textbf{130} Tree Sparrow (land) & \textbf{131} Vesper Sparrow (land) & \textbf{132} White crowned Sparrow (land) \\
\textbf{133} White throated Sparrow (land) & \textbf{134} Cape Glossy Starling (land) & \textbf{135} Bank Swallow (land) & \textbf{136} Barn Swallow (land) \\
\textbf{137} Cliff Swallow (land) & \textbf{138} Tree Swallow (land) & \textbf{139} Scarlet Tanager (land) & \textbf{140} Summer Tanager (land) \\
\textbf{141} Artic Tern (water) & \textbf{142} Black Tern (water) & \textbf{143} Caspian Tern (water) & \textbf{144} Common Tern (water) \\
\textbf{145} Elegant Tern (water) & \textbf{146} Forsters Tern (water) & \textbf{147} Least Tern (water) & \textbf{148} Green tailed Towhee (land) \\
\textbf{149} Brown Thrasher (land) & \textbf{150} Sage Thrasher (land) & \textbf{151} Black capped Vireo (land) & \textbf{152} Blue headed Vireo (land) \\
\textbf{153} Philadelphia Vireo (land) & \textbf{154} Red eyed Vireo (land) & \textbf{155} Warbling Vireo (land) & \textbf{156} White eyed Vireo (land) \\
\textbf{157} Yellow throated Vireo (land) & \textbf{158} Bay breasted Warbler (land) & \textbf{159} Black and white Warbler (land) & \textbf{160} Black throated Blue Warbler (land) \\
\textbf{161} Blue winged Warbler (land) & \textbf{162} Canada Warbler (land) & \textbf{163} Cape May Warbler (land) & \textbf{164} Cerulean Warbler (land) \\
\textbf{165} Chestnut sided Warbler (land) & \textbf{166} Golden winged Warbler (land) & \textbf{167} Hooded Warbler (land) & \textbf{168} Kentucky Warbler (land) \\
\textbf{169} Magnolia Warbler (land) & \textbf{170} Mourning Warbler (land) & \textbf{171} Myrtle Warbler (land) & \textbf{172} Nashville Warbler (land) \\
\textbf{173} Orange crowned Warbler (land) & \textbf{174} Palm Warbler (land) & \textbf{175} Pine Warbler (land) & \textbf{176} Prairie Warbler (land) \\
\textbf{177} Prothonotary Warbler (land) & \textbf{178} Swainson Warbler (land) & \textbf{179} Tennessee Warbler (land) & \textbf{180} Wilson Warbler (land) \\
\textbf{181} Worm eating Warbler (land) & \textbf{182} Yellow Warbler (land) & \textbf{183} Northern Waterthrush (land) & \textbf{184} Louisiana Waterthrush (land) \\
\textbf{185} Bohemian Waxwing (land) & \textbf{186} Cedar Waxwing (land) & \textbf{187} American Three toed Woodpecker (land) & \textbf{188} Pileated Woodpecker (land) \\
\textbf{189} Red bellied Woodpecker (land) & \textbf{190} Red cockaded Woodpecker (land) & \textbf{191} Red headed Woodpecker (land) & \textbf{192} Downy Woodpecker (land) \\
\textbf{193} Bewick Wren (land) & \textbf{194} Cactus Wren (land) & \textbf{195} Carolina Wren (land) & \textbf{196} House Wren (land) \\
\textbf{197} Marsh Wren (land) & \textbf{198} Rock Wren (land) & \textbf{199} Winter Wren (land) & \textbf{200} Common Yellowthroat (land) \\
\end{longtable}
\renewcommand{\arraystretch}{1.0}
\normalsize
\clearpage
\twocolumn

\begin{table*}[t]
\centering
\scriptsize
\setlength{\tabcolsep}{1.2pt}
\resizebox{\textwidth}{!}{%
\begin{tabular}{ll*{15}{c}}
\toprule
& & \multicolumn{5}{c}{EVQA} & \multicolumn{5}{c}{EIC} & \multicolumn{5}{c}{VLKEB} \\
\cmidrule(lr){3-7}\cmidrule(lr){8-12}\cmidrule(lr){13-17}
\textbf{\# Edits} & \textbf{Method} & Rel & T-Gen & I-Gen & T-Loc & I-Loc & Rel & T-Gen & I-Gen & T-Loc & I-Loc & Rel & T-Gen & I-Gen & T-Loc & I-Loc \\
\midrule
\multirow{6}{*}{1} & FT-VL & 100.00 & 91.03 & 95.38 & 95.29 & 90.53 & \textbf{99.93} & \textbf{94.65} & \textbf{97.34} & 91.63 & 82.16 & \textbf{100.00} & \textbf{98.74} & \textbf{99.92} & 76.13 & 84.15 \\
 & TP & 33.27 & 31.33 & 29.06 & 96.77 & 98.83 & 57.49 & 57.48 & 53.91 & 95.41 & 96.52 & 50.34 & 52.43 & 50.27 & 92.59 & 96.58 \\
 & MEND & 91.23 & 90.05 & 91.29 & 91.02 & 90.22 & 92.82 & 91.81 & 90.59 & 96.38 & 93.69 & 92.13 & 91.28 & 90.22 & 89.19 & 90.13 \\
 & SERAC & 89.33 & 83.72 & 84.97 & 82.05 & 23.78 & 88.18 & 81.03 & 85.61 & 84.01 & 28.58 & 89.77 & 89.11 & 87.92 & 66.68 & 14.20 \\
 & LiveEdit & 94.28 & \textbf{94.51} & 88.01 & 100.00 & 100.00 & 82.16 & 81.01 & 78.27 & 100.00 & 100.00 & 96.43 & 95.22 & 93.72 & 100.00 & \textbf{100.00} \\
 & \textbf{VICE} & \textbf{100.00} & 87.33 & \textbf{97.57} & \textbf{100.00} & \textbf{100.00} & 98.91 & 92.76 & 79.61 & \textbf{100.00} & \textbf{100.00} & 99.98 & 98.27 & 96.72 & \textbf{100.00} & 99.98 \\
\cmidrule(lr){1-17}
\multirow{6}{*}{10} & FT-VL & 91.05 & 81.96 & 85.11 & 83.89 & 68.17 & 85.74 & 81.72 & \textbf{79.73} & 77.89 & 77.60 & 86.31 & 82.23 & 84.95 & 66.64 & 80.62 \\
 & TP & 32.98 & 30.90 & 28.77 & 91.08 & 96.14 & 55.10 & 55.04 & 51.65 & 75.19 & 88.18 & 49.59 & 52.15 & 49.70 & 87.02 & 93.70 \\
 & MEND & 3.58 & 3.55 & 3.53 & 2.10 & 1.26 & 66.49 & 64.77 & 58.42 & 87.25 & 86.15 & 0.18 & 0.24 & 0.05 & 0.03 & 0.19 \\
 & SERAC & 88.09 & 83.40 & 83.57 & 64.91 & 15.50 & 56.57 & 56.88 & 52.62 & 59.96 & 14.66 & 81.55 & 74.49 & 80.24 & 54.71 & 13.15 \\
 & LiveEdit & 93.79 & \textbf{93.21} & 86.42 & 100.00 & 100.00 & 81.93 & 80.80 & 75.55 & 100.00 & 100.00 & 95.54 & 94.52 & 91.25 & 100.00 & \textbf{100.00} \\
 & \textbf{VICE} & \textbf{100.00} & 87.18 & \textbf{97.53} & \textbf{100.00} & \textbf{100.00} & \textbf{98.91} & \textbf{92.76} & 77.92 & \textbf{100.00} & \textbf{100.00} & \textbf{99.98} & \textbf{98.27} & \textbf{97.14} & \textbf{100.00} & 99.96 \\
\cmidrule(lr){1-17}
\multirow{6}{*}{100} & FT-VL & 76.44 & 67.30 & 67.76 & 58.04 & 46.18 & 73.11 & 69.78 & 65.46 & 52.21 & 58.69 & 75.91 & 72.95 & 74.89 & 56.36 & 70.29 \\
 & TP & 32.60 & 30.74 & 28.65 & 83.13 & 89.68 & 50.31 & 50.67 & 46.83 & 34.35 & 63.78 & 48.26 & 50.78 & 48.20 & 64.18 & 79.15 \\
 & MEND & 2.22 & 2.20 & 2.21 & 0.21 & 0.62 & 56.83 & 56.89 & 53.07 & 87.63 & 84.64 & 0.56 & 0.58 & 0.66 & 0.18 & 0.07 \\
 & SERAC & 88.08 & 81.53 & 82.48 & 62.13 & 12.90 & 53.35 & 53.70 & 49.41 & 48.04 & 17.25 & 72.25 & 62.43 & 70.68 & 53.73 & 13.69 \\
 & LiveEdit & 93.54 & \textbf{92.34} & 85.89 & 100.00 & 99.31 & 80.81 & 78.77 & 63.52 & 100.00 & \textbf{100.00} & 94.56 & 90.65 & 89.56 & 100.00 & \textbf{100.00} \\
 & \textbf{VICE} & \textbf{100.00} & 87.15 & \textbf{97.50} & \textbf{100.00} & \textbf{99.82} & \textbf{98.91} & \textbf{92.76} & \textbf{73.97} & \textbf{100.00} & 99.95 & \textbf{99.98} & \textbf{98.27} & \textbf{97.14} & \textbf{100.00} & 99.85 \\
\cmidrule(lr){1-17}
\multirow{6}{*}{1000} & FT-VL & 60.21 & 50.76 & 51.03 & 24.37 & 29.48 & 64.21 & 59.43 & 57.09 & 33.39 & 38.30 & 65.30 & 62.67 & 63.87 & 33.89 & 37.32 \\
 & TP & 32.82 & 30.48 & 28.55 & 42.12 & 60.28 & 33.75 & 36.27 & 32.34 & 10.47 & 26.22 & 34.23 & 34.74 & 34.44 & 27.11 & 45.52 \\
 & MEND & 0.04 & 0.05 & 0.05 & 0.08 & 0.09 & 54.39 & 54.14 & 50.99 & 83.87 & 80.60 & 0.03 & 0.05 & 0.07 & 0.06 & 0.08 \\
 & SERAC & 85.57 & 75.58 & 82.01 & 62.46 & 15.69 & 52.93 & 53.44 & 49.01 & 49.91 & 16.65 & 60.93 & 56.49 & 60.06 & 52.94 & 15.04 \\
 & LiveEdit & 92.93 & \textbf{90.16} & 84.30 & 100.00 & 96.43 & 72.80 & 69.95 & 57.05 & 100.00 & 99.79 & 92.22 & 83.97 & 82.75 & 100.00 & \textbf{100.00} \\
 & \textbf{VICE} & \textbf{100.00} & 87.03 & \textbf{97.01} & \textbf{100.00} & \textbf{98.38} & \textbf{98.91} & \textbf{92.76} & \textbf{60.91} & \textbf{100.00} & \textbf{99.90} & \textbf{99.98} & \textbf{98.26} & \textbf{96.75} & \textbf{100.00} & 98.32 \\
\bottomrule
\end{tabular}%
}
\caption{Complete lifelong editing results for LLaVA-v1.5 on EVQA, EIC, and VLKEB. FT-VL, TP, and \vice{} are from our local runs; MEND, SERAC, and LiveEdit entries are copied from \citet{chen2025liveedit}.}
\label{tab:appendix_llava_results}
\end{table*}

\begin{table*}[t]
\centering
\scriptsize
\setlength{\tabcolsep}{1.2pt}
\resizebox{\textwidth}{!}{%
\begin{tabular}{ll*{15}{c}}
\toprule
& & \multicolumn{5}{c}{EVQA} & \multicolumn{5}{c}{EIC} & \multicolumn{5}{c}{VLKEB} \\
\cmidrule(lr){3-7}\cmidrule(lr){8-12}\cmidrule(lr){13-17}
\textbf{\# Edits} & \textbf{Method} & Rel & T-Gen & I-Gen & T-Loc & I-Loc & Rel & T-Gen & I-Gen & T-Loc & I-Loc & Rel & T-Gen & I-Gen & T-Loc & I-Loc \\
\midrule
\multirow{6}{*}{1} & FT-VL & \textbf{100.00} & \textbf{99.74} & \textbf{99.86} & 63.66 & 35.65 & \textbf{100.00} & \textbf{95.28} & \textbf{96.41} & 58.21 & 33.84 & \textbf{100.00} & \textbf{99.75} & \textbf{99.95} & 56.12 & 42.03 \\
 & TP & 30.07 & 28.55 & 30.40 & 89.48 & 71.50 & 30.79 & 32.00 & 34.41 & 44.72 & 28.74 & 37.17 & 38.31 & 37.49 & 74.56 & 46.15 \\
 & MEND & 93.47 & 91.32 & 91.22 & 81.51 & 74.97 & 90.28 & 83.91 & 79.05 & 95.92 & 93.97 & 90.28 & 89.18 & 89.55 & 93.63 & 86.69 \\
 & SERAC & 87.66 & 69.24 & 85.96 & 82.11 & 24.11 & 85.09 & 78.96 & 81.70 & 83.36 & 24.50 & 89.28 & 89.06 & 87.35 & 64.88 & 12.72 \\
 & LiveEdit & 94.90 & 92.63 & 92.06 & 100.00 & 100.00 & 86.84 & 83.82 & 76.95 & 100.00 & 100.00 & 96.41 & 95.56 & 92.85 & 100.00 & \textbf{100.00} \\
 & \textbf{VICE} & 99.98 & 88.74 & 97.35 & \textbf{100.00} & \textbf{100.00} & 98.93 & 93.67 & 77.85 & \textbf{100.00} & \textbf{100.00} & 95.27 & 82.95 & 92.05 & \textbf{100.00} & 99.99 \\
\cmidrule(lr){1-17}
\multirow{6}{*}{10} & FT-VL & 72.26 & 64.96 & 67.67 & 56.13 & 30.94 & 73.47 & 70.26 & 67.19 & 29.40 & 14.59 & 75.21 & 73.90 & 73.35 & 48.01 & 35.70 \\
 & TP & 18.82 & 18.45 & 18.58 & 38.24 & 21.89 & 14.54 & 14.61 & 14.49 & 2.78 & 1.25 & 14.66 & 14.60 & 13.90 & 9.60 & 3.89 \\
 & MEND & 37.34 & 31.66 & 35.52 & 44.31 & 37.24 & 69.98 & 69.35 & 59.25 & 93.39 & 91.52 & 57.01 & 55.39 & 55.34 & 72.03 & 67.66 \\
 & SERAC & 86.61 & 67.86 & 83.14 & 67.69 & 14.47 & 50.72 & 49.60 & 48.36 & 62.87 & 16.01 & 79.90 & 71.80 & 78.44 & 52.64 & 13.08 \\
 & LiveEdit & 94.06 & \textbf{91.84} & 90.60 & 100.00 & 99.92 & 85.79 & 83.52 & \textbf{76.09} & 100.00 & 100.00 & \textbf{95.75} & \textbf{95.13} & 91.99 & 100.00 & \textbf{100.00} \\
 & \textbf{VICE} & \textbf{99.98} & 88.54 & \textbf{97.31} & \textbf{100.00} & \textbf{100.00} & \textbf{98.93} & \textbf{93.67} & 76.08 & \textbf{100.00} & \textbf{100.00} & 95.27 & 82.95 & \textbf{92.44} & \textbf{100.00} & 99.97 \\
\cmidrule(lr){1-17}
\multirow{6}{*}{100} & FT-VL & 56.02 & 49.77 & 50.35 & 16.97 & 10.42 & 59.24 & 56.32 & 52.58 & 4.13 & 4.65 & 64.49 & 63.10 & 62.70 & 21.95 & 14.10 \\
 & TP & 6.64 & 6.16 & 6.04 & 1.80 & 0.61 & 7.74 & 7.44 & 7.23 & 1.08 & 0.63 & 4.93 & 4.85 & 4.40 & 1.57 & 1.34 \\
 & MEND & 23.32 & 20.75 & 21.42 & 47.27 & 42.60 & 51.92 & 53.96 & 48.24 & 86.02 & 83.88 & 38.85 & 40.02 & 37.99 & 72.82 & 68.89 \\
 & SERAC & 87.39 & 66.53 & 81.70 & 65.13 & 14.04 & 47.41 & 46.17 & 45.22 & 49.37 & 13.52 & 69.38 & 58.31 & 67.57 & 51.71 & 13.72 \\
 & LiveEdit & 93.23 & \textbf{91.27} & 88.34 & 100.00 & 99.61 & 84.38 & 83.53 & \textbf{71.85} & 100.00 & \textbf{100.00} & 94.06 & \textbf{94.62} & 89.42 & 100.00 & \textbf{100.00} \\
 & \textbf{VICE} & \textbf{99.98} & 88.49 & \textbf{97.28} & \textbf{100.00} & \textbf{99.82} & \textbf{98.93} & \textbf{93.67} & 71.73 & \textbf{100.00} & 99.95 & \textbf{95.27} & 82.95 & \textbf{92.44} & \textbf{100.00} & 99.87 \\
\cmidrule(lr){1-17}
\multirow{6}{*}{1000} & FT-VL & 36.32 & 31.16 & 34.89 & 5.27 & 5.86 & 62.25 & 57.95 & 54.73 & 7.16 & 11.06 & 42.64 & 41.62 & 41.18 & 8.55 & 9.32 \\
 & TP & 7.70 & 6.99 & 7.50 & 1.70 & 0.55 & 8.16 & 8.26 & 7.63 & 0.59 & 1.46 & 3.74 & 3.47 & 3.57 & 1.17 & 2.75 \\
 & MEND & 31.84 & 26.60 & 33.98 & 43.08 & 44.90 & 49.30 & 51.46 & 46.70 & 81.03 & 82.53 & 42.35 & 45.07 & 42.68 & 66.98 & 62.32 \\
 & SERAC & 84.50 & 60.36 & 81.83 & 63.26 & 13.40 & 47.03 & 45.79 & 44.82 & 42.12 & 15.31 & 57.02 & 51.66 & 56.80 & 48.48 & 13.24 \\
 & LiveEdit & 92.75 & \textbf{89.81} & 85.12 & 100.00 & 96.82 & 81.43 & 80.34 & \textbf{62.25} & 100.00 & 99.73 & 93.41 & \textbf{89.87} & 85.21 & 100.00 & \textbf{100.00} \\
 & \textbf{VICE} & \textbf{99.98} & 88.37 & \textbf{96.82} & \textbf{100.00} & \textbf{98.26} & \textbf{98.93} & \textbf{93.67} & 57.31 & \textbf{100.00} & \textbf{99.92} & \textbf{95.27} & 82.95 & \textbf{92.14} & \textbf{100.00} & 98.29 \\
\bottomrule
\end{tabular}%
}
\caption{Complete lifelong editing results for MiniGPT4 on EVQA, EIC, and VLKEB. FT-VL, TP, and \vice{} are from our local runs; MEND, SERAC, and LiveEdit entries are copied from \citet{chen2025liveedit}.}
\label{tab:appendix_minigpt_results}
\end{table*}

\end{document}